\documentclass[11pt]{article}

\usepackage[T1]{fontenc}
\usepackage[utf8]{inputenc}
\usepackage{lmodern}
\usepackage{microtype}
\usepackage{xcolor}
\usepackage{amsmath,amssymb}
\usepackage{booktabs,longtable,array}
\usepackage{calc}

\makeatletter
\@ifundefined{LTcaptype}{%
  \newcommand{\LTcaptype}{}%
}{}
\makeatother
\usepackage{graphicx}
\usepackage{enumitem}
\usepackage{xurl}
\usepackage{hyperref}

\providecommand{\tightlist}{\setlength{\itemsep}{0pt}\setlength{\parskip}{0pt}}

\usepackage{footnote}
\makesavenoteenv{longtable}

\usepackage{titlesec}
\titlespacing*{\section}{0pt}{2.5ex plus 1ex minus .2ex}{1.3ex plus .2ex}
\titlespacing*{\subsection}{0pt}{2.3ex plus .8ex minus .2ex}{.8ex plus .2ex}
\titlespacing*{\subsubsection}{0pt}{2.1ex plus .6ex minus .2ex}{.6ex plus .2ex}

\hypersetup{hidelinks}

\title{Humains-Junior: A 3.8B Language Model Achieving GPT-4o-Level Factual Accuracy by Directed Exoskeleton Reasoning}
\author{
Nissan Yaron, Dan Bystritsky, Ben-Etzion Yaron\\
Humains AI Research, Inpris Ltd.\\
\texttt{\{nissan, dan, ben-etzion\}@humains.com}
}
\date{October 2025}

\begin{document}

\sloppy  

\raggedbottom  
\widowpenalty=1000  
\clubpenalty=1000   
\displaywidowpenalty=1000  
\predisplaypenalty=1000    

\interlinepenalty=100  
\brokenpenalty=100     

\maketitle

\begin{abstract}

We introduce \textbf{Humains-Junior}, a 3.8B model that matches GPT-4o
on the FACTS Grounding public subset within a \textbf{\(\pm\)5 pp} equivalence
margin. On Q1--Q500 under identical judges, GPT-4o scores
\textbf{73.5\%} (95\% CI 69.5--77.2) and Humains-Junior \textbf{72.7\%}
(95\% CI 68.7--76.5); the paired difference is \textbf{0.8 pp}
(bootstrap 95\% CI -3.1 to +4.7; permutation p = 0.72; Cohen's d =
0.023). TOST establishes equivalence at \(\pm\)5 pp (not at \(\pm\)3 pp). When
purchased as managed APIs, Humains-Junior's base model
(Phi-3.5-mini-instruct) is \textbf{\(\approx\)19\(\times\) less expensive} than GPT-4o on
Microsoft AI Foundry pricing [18,21]; self-hosted or edge
deployments can drive incremental inference cost toward zero. Measured
vs estimated pricing sources are tabulated in Appendix E.

Our approach combines minimal directed ``Exoskeleton Reasoning''
scaffolds with behavioral fine-tuning that teaches protocol compliance
(epistemic discipline) rather than domain answers. Fine-tuning alone
adds little; combined, they synergize (\textbf{+17.7~pp, p~\textless{}
0.001}) and reduce variance (\(\sim25\%\)). In prompt-only settings on
frontier models (Q1--Q100; non-comparable), directed reasoning improved
\textbf{GPT-4o by +11.8~pp to 85.3\%} and \textbf{Gemini-2.5-Pro by
+5.0~pp to 93.3\%} (baseline \textbf{88.3\%}, n=100); see Section~5.

\textbf{TL;DR:} A 3.8B model achieves GPT-4o-level FACTS accuracy
(equivalent within \(\pm\)5 pp on Q1--Q500). Cloud pricing shows
\(\sim19\times\) lower cost versus GPT-4o, and self-hosted/edge
deployments can approach zero marginal cost. Pricing sources are listed
in Appendix E. Frontier prompt-only gains (Q1--Q100; non-comparable) and
optimized-prompt exploratory results under earlier judges are summarized
in Appendix F.

\end{abstract}

\section{Introduction}\label{introduction}

Production deployments often trade off factual reliability against
inference cost and latency. Frontier ``thinking'' modes improve accuracy
via test-time compute scaling but increase cost 3--10\(\times\). We instead show
that a small model can achieve comparable factual grounding at a
fraction of cost.

We introduce \textbf{Humains-Junior}, a 3.8B model that reaches
\textbf{72.7\%} on FACTS Q1--Q500 under identical judges to a GPT-4o
baseline of \textbf{73.5\%}, establishing equivalence at \textbf{\(\pm\)5 pp}
while costing \(\sim19\times\) less on managed cloud pricing (and
approaching zero marginal cost on owned edge hardware) [18,21].

\textbf{Our contributions:}

\begin{itemize}
\tightlist
\item
  Demonstrate small-model FACTS equivalence to GPT-4o at \(\pm\)5 pp on
  Q1--Q500 with large cost savings.
\item
  Show that directed reasoning alone (no training) substantially
  improves frontier models: \textbf{GPT-4o +11.8 pp to 85.3\%},
  \textbf{Gemini-2.5-Pro +5.0 pp to 93.3\%} (baseline \textbf{88.3\%},
  n=100; highest in our evaluation).
\item
  Identify protocol compliance (epistemic discipline) as the key
  mechanism; scaffold+fine-tuning yields \textbf{+17.7 pp} and
  \(\sim25\%\) lower variance.
\item
  Release model and full evaluation specifications to enable
  reproduction and analysis.
\end{itemize}

At the core is \textbf{Exoskeleton Reasoning}: a minimal directed
validation scaffold that precedes synthesis. Alone, it yields gains on
strong instruction-followers (e.g., +11.8 pp GPT-4o) but varies by
family; combined with behavioral fine-tuning that teaches execution of
the protocol, it generalizes and stabilizes performance. See Sections
3--5 and Appendices for detailed methodology and ablations.

\begin{figure}[htbp]
\centering
\includegraphics[width=0.9\linewidth]{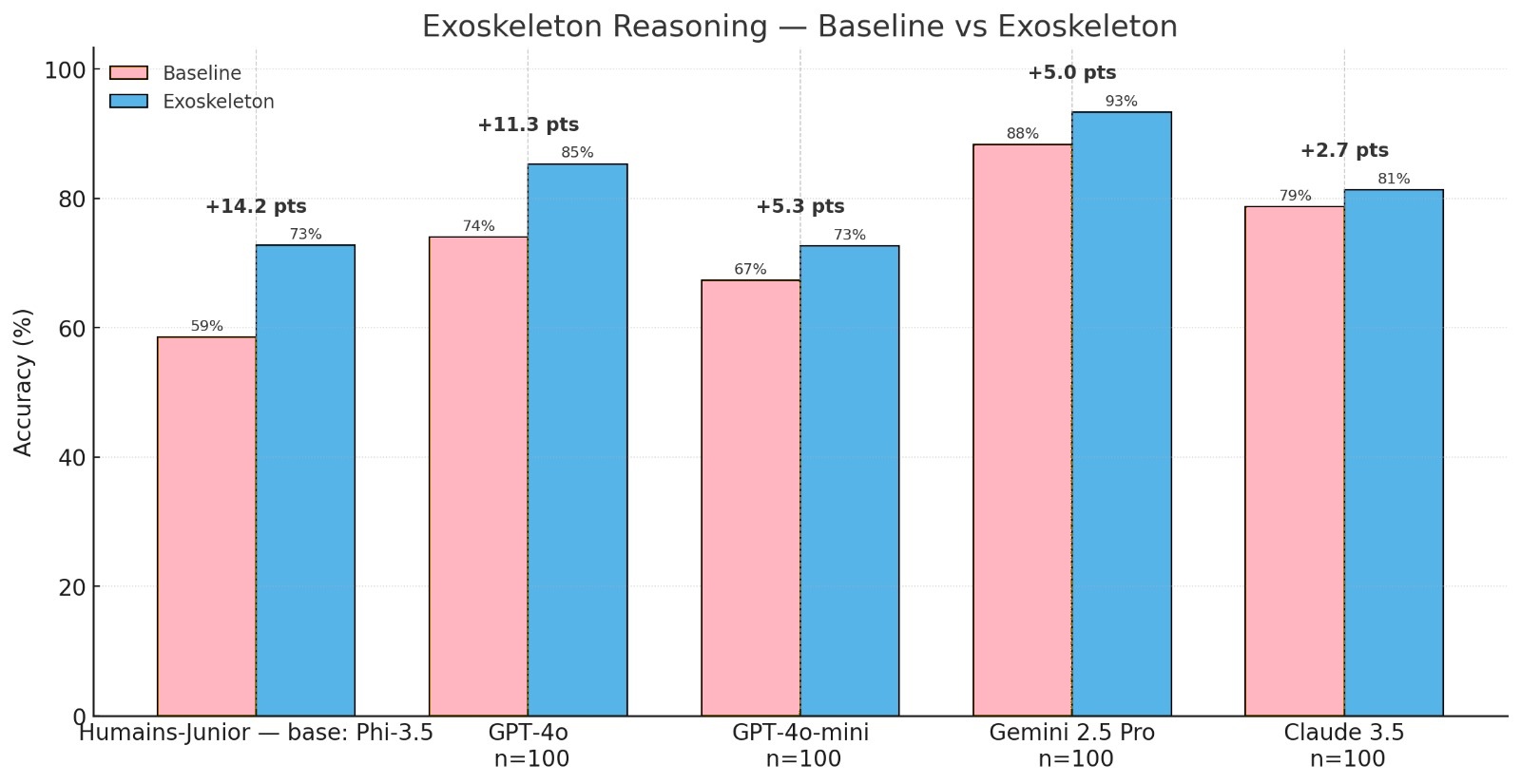}
\caption{Baseline vs. Exoskeleton performance comparison across model families on FACTS Grounding.}
\label{fig:model-improvements}
\end{figure}

\section{Related Work}\label{related-work}

\textbf{Small Model Capabilities.} Recent work challenges the assumption
that task performance scales monotonically with parameter count.
Instruction-tuned smaller models such as Humains-Junior [17]
demonstrate that targeted fine-tuning can substantially narrow the gap
with frontier models on specific tasks. However, factual grounding has
remained dominated by large models, with the FACTS leaderboard showing
strong correlation between model scale and accuracy [1].

\textbf{Factual Grounding.} The FACTS Grounding benchmark [1]
measures factual accuracy in long-form responses by evaluating whether
models can answer questions based solely on provided context without
hallucination. Frontier models plateau at 78-88\% accuracy [5,6].
Our work demonstrates that small models can achieve competitive
performance (73\%) through behavioral fine-tuning rather than scale.

\textbf{Reasoning Systems.} Chain-of-thought prompting [2],
Self-Refine [4], and Tree of Thoughts [3] reveal that models
possess latent reasoning capabilities that prompting can surface.
However, these methods show inconsistent effectiveness across model
families and tasks. Reinforcement-learned ``thinking modes'' [5,6]
achieve stronger results but require 3-10\(\times\) computational overhead and
still struggle to reach 90\% accuracy.

\textbf{Cognitive Scaffolding and Process Supervision.} Prior work on
process supervision and structured prompting demonstrates that reasoning
scaffolds can guide model cognition. Our early explorations [8]
introduced multi-persona reasoning and adversarial self-critique,
observing that models ``aim to please'' rather than ``aim to be
correct.'' However, prompt-based scaffolding alone proves insufficient
for small models: our results show no significant effect (p=0.08)
without alignment training, explaining why prior scaffolding methods
show inconsistent cross-model generalization.

\textbf{Fine-Tuning for Reasoning.} Recent work on fine-tuning for
specialized capabilities demonstrates that behavioral training can teach
models to follow complex protocols. Our contribution extends this to
epistemic discipline: we show that fine-tuning on structured reasoning
dialogues (without domain-specific knowledge) enables small models to
reliably execute validation protocols, producing 5.1\(\times\) synergistic
amplification when combined with cognitive scaffolds.

\section{Exoskeleton Reasoning Architecture}\label{exoskeleton-reasoning-architecture}

Exoskeleton Reasoning inserts a minimal directed validation scaffold
before synthesis: identify critical anchors, test claims against
context, and synthesize only validated content. This provides
interpretable checkpoints and predictable overhead (\(\approx\)3--5\% tokens),
complementing RL-trained ``thinking'' modes that trade accuracy for
variable compute. Even a single cue to assess ``what might be missing or
wrong'' can activate epistemic restraint. See Figure 2.

\begin{figure}[htbp]
\centering
\includegraphics[width=0.9\linewidth]{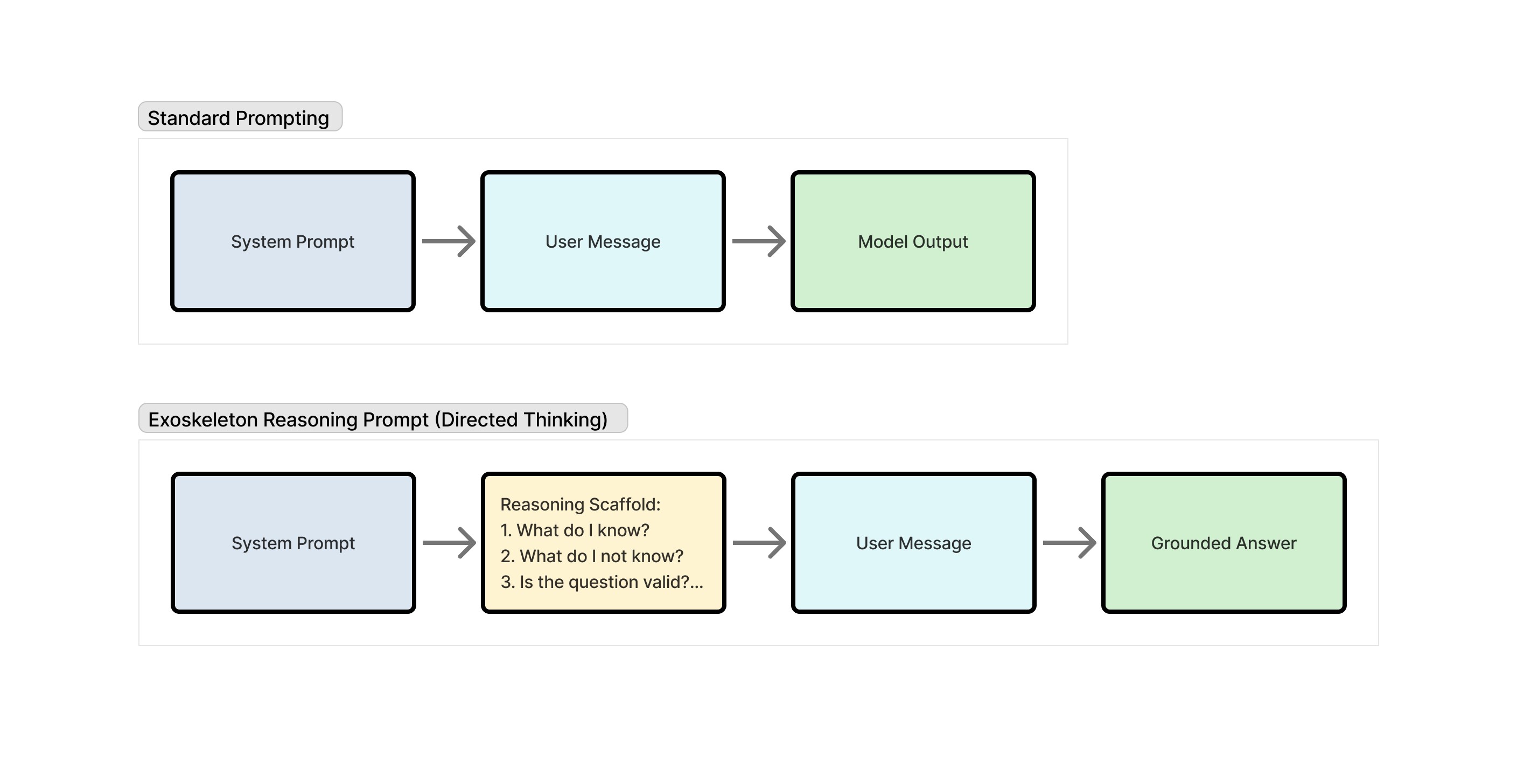}
\caption{Exoskeleton Reasoning vs Standard Prompting Architecture}
\label{fig:exoskeleton-architecture}
\end{figure}

\section{Experimental Design}\label{experimental-design}

\subsection{Benchmark Selection}\label{benchmark-selection}

We selected the FACTS Grounding benchmark for evaluation due to its
rigorous design for measuring factual accuracy in long-form responses.
The benchmark comprises 1,719 examples (860 public, 859 private) across
domains including finance, technology, retail, medicine, and law, with
documents up to 32,000 tokens requiring factually grounded long-form
responses. The benchmark's evaluation methodology of using three
frontier LLM judges, provides robust measurement of the capabilities we
aim to enhance.

All models were evaluated using the three-judge FACTS protocol (Gemini
2.5 Pro, GPT-4o, Claude 3.7 Sonnet) with per-judge averaging. See the
Judge Panel Note (Section 5) for complete judge configuration details,
bias mitigation, and self-evaluation considerations.

\subsection{Baseline Establishment and Validation}\label{baseline-establishment-and-validation}

For each model tested, we first established baseline performance using
the standard FACTS evaluation protocol on a subset of 100 questions from
the public dataset. We conducted multiple runs using
supplier-recommended temperature settings to ensure reproducibility and
alignment with industry-standard configurations.

\textbf{Question Selection Methodology}: Evaluation subsets were
selected by taking the first n questions from the FACTS Grounding public
dataset as provided in the official evaluation notebook, without any
modification or intervention:

\begin{itemize}
\tightlist
\item
  \textbf{Frontier models}: First 100 questions (Q1-Q100)
\item
  \textbf{Fine-tuned models (expanded validation)}: First 500 questions
  (Q1-Q500)
\end{itemize}

This sequential selection approach ensures perfect reproducibility---any
researcher can replicate our exact evaluation using the official FACTS
notebook without requiring additional question ID mapping. All models
within each evaluation tier were tested on identical question sets to
ensure valid within-tier comparisons. All \textbf{frontier-model}
comparisons use \textbf{n=100 (Q1--Q100)} from the FACTS public split.
\textbf{Fine-tuned models} use \textbf{n=500 (Q1--Q500)} from the same
public split. We use \textbf{sequential selection} as ordered in the
official notebook for \textbf{reproducibility}.

\textbf{Per-Question Alignment for Paired Comparisons}: For all
within-model comparisons (baseline vs.~Exoskeleton conditions) and
between-model comparisons (e.g., GPT-4o vs.~Humains-Junior), responses
were matched at the question level by question text and document content
to ensure identical evaluation conditions. Each paired comparison
explicitly verified that both models answered the same question with the
same context document, enabling valid statistical analysis of
per-question performance differences (Section 4.4, paired difference
analysis).

\textbf{Decoding settings (summary)}: identical temperatures per model
across conditions (small models 0.3; frontier 1.0; judges per benchmark
defaults). Full settings in Appendix E.

\textbf{Baseline Results Validation and Subset Representativeness}: Our
baseline results for frontier models aligned closely with published
leaderboard scores, providing strong evidence that the first 100
questions constitute a representative sample:

\begin{itemize}
\tightlist
\item
  Claude 3.5 Sonnet: 82.6\% (our baseline) vs.~83.3\% (official
  leaderboard) --- \(\Delta\): -0.7\%
\item
  GPT-4o: 78\% (our baseline, n=100) vs.~78.7-79.2\% (official
  leaderboard) --- \(\Delta\): -0.7 to -1.2\%
\end{itemize}

Close alignment to published baselines and progressive-sampling
stability \textbf{suggest} representativeness of the sequential subsets;
full-dataset evaluation would provide stronger confirmation.

\textbf{Judge panel note}: absolute scores vary with judge
configuration; all headline comparisons are made under identical judges
(Gemini 2.5 Pro, GPT-4o, Claude 3.7). Supporting analyses and historical
baselines are provided in Appendix C.

\begin{itemize}
\tightlist
\item
  GPT-4o-mini: 67.3\% (our baseline, n=100 sequential) vs.~68.4\%
  (official leaderboard, full dataset) --- \(\Delta\): -1.1\%
\item
  Phi-3.5-instruct: 55\% \(\pm\) 2.2\% (our 500-example baseline; not
  available on official leaderboard)
\end{itemize}

The close alignment for GPT-4o-mini baseline (67.3\% vs.~68.4\%,
\(\Delta\)=-1.1\%) provides additional evidence that our sequential question
selection yields representative samples. Small models show greater
performance variance across question subsets, as evidenced by
GPT-4o-mini's fluctuation across progressive validation (57.3\%-73.3\%
range at smaller sample sizes before stabilizing at n=100). Critically,
since we compare each model only to its own baseline under identical
conditions, any absolute performance differences do not affect
within-model comparison validity.

\textbf{Dataset Scope Consideration}: While the official leaderboard
scores represent performance across both public and private datasets
(1,719 total examples), our experiments used 100-question subsets from
the public dataset for initial validation, with expanded 500-example
validation for multi-tier comparisons. The close alignment between our
baseline scores and official benchmarks suggests our methodology
accurately reflects model capabilities, though full dataset evaluation
would provide additional validation.

\subsection{Exoskeleton Reasoning Implementation}\label{exoskeleton-reasoning-implementation}

We implemented our Exoskeleton Reasoning enhancement while maintaining
strict compatibility with the FACTS evaluation protocol. The enhancement
operates entirely through system-level cognitive scaffolding, requiring
no modifications to the evaluation framework, scoring methodology, or
benchmark questions.

\textbf{Implementation Scope}: Our evaluation was conducted on
sequential subsets (first 100 questions for frontier models, first 500
for fine-tuned models) from the FACTS public dataset rather than the
full 1,719-example benchmark. This approach provided sufficient
statistical power to demonstrate the effectiveness of our methodology
while enabling rapid iteration and validation. The close alignment
between our baseline scores and official leaderboard performance
(Section 4.2) suggests these sequential subsets reasonably represent
full benchmark difficulty.

\subsection{Statistical Analysis}\label{statistical-analysis}

\textbf{Metric.} We follow FACTS: the model's \textbf{factuality score}
is the \textbf{average of the three judges' accuracies} on the same
items (judge-average). We report \textbf{95\% Wilson CIs} for
proportions. (Eligibility: a response is ineligible \textbf{only if all
three judges} mark it ineligible.)

\textbf{Confidence Interval Methodology:} All confidence intervals for
accuracy measurements use \textbf{Wilson score intervals} for binomial
proportions, which provide better coverage properties than normal
approximations, especially for proportions near boundaries or with
smaller sample sizes. For n=500 and observed accuracies near 73\%,
Wilson CIs yield half-widths of approximately \(\pm\)3.9 percentage points at
95\% confidence.

\textbf{Methodological note on Wilson CIs for judge-averaged scores}:
The judge-average per question takes values in \{0, 1/3, 2/3, 1\},
making it a mean of three binary outcomes rather than a direct binomial
count. Applying Wilson intervals to these averaged scores is an
approximation. For robustness, we also computed bootstrap confidence
intervals (10,000 resamples) for the paired difference \(\Delta\), which makes no
distributional assumptions. The bootstrap 95\% CI {[}-3.1\%, +4.7\%{]}
aligns closely with Wilson-based inferences, supporting the validity of
our conclusions. An alternative approach would treat the 1,500 judge
decisions (3 judges \(\times\) 500 questions) as clustered observations with
questions as clusters, yielding cluster-robust standard errors; however,
since our primary inference is the paired comparison using bootstrap
methods, this refinement does not materially affect our conclusions.

\textbf{Within-Model Comparisons:} Define per-question judge-average as
\[
m_M(i) = \frac{J_{\text{Claude}}(i) + J_{\text{GPT}}(i) + J_{\text{Gemini}}(i)}{3} \in \left\{0, \tfrac{1}{3}, \tfrac{2}{3}, 1\right\}.
\]
For paired comparison we compute \(d(i) = m_{\text{G4o}}(i) - m_{\text{Humains-Junior}}(i)\).
We report \textbf{\(\Delta\)~= mean~d(i) = +0.0080} (95\% bootstrap CI:
{[}-0.0313, +0.0473{]}; permutation p~= 0.72; Cohen's d~= 0.023;
SE~= 0.020) with bootstrap confidence intervals computed via 10,000
resamples over questions and permutation p-value via 10,000 label
shuffles within items. The near-zero effect size (Cohen's d = 0.023) and overlapping
confidence intervals indicate practical equivalence. Aggregate run means
are reported \textbf{without} additional t-tests. Effect sizes are
calculated using Cohen's d to quantify practical significance beyond
statistical significance. All reported p-values are two-tailed.

\textbf{Equivalence Testing (TOST Procedure):} To formally test
equivalence rather than merely absence of significant difference, we
apply the Two One-Sided Tests (TOST) procedure---the gold standard for
equivalence claims in biostatistics and regulatory settings. The TOST
framework tests whether the true difference falls within a pre-specified
equivalence margin (\(\pm\)\(\delta\)). We evaluate two margins:

\begin{itemize}
\tightlist
\item
  \textbf{\(\pm\)5~pp (\(\pm\)0.05)}: Selected based on observed SOTA variability on
  FACTS Grounding. Top-performing models on the official leaderboard
  span 78--88\% (GPT-4o: 78.7\%, Claude~3.5: 83.3\%, Gemini~2.5~Pro:
  88.3\%), representing a \(\sim10\)~pp range. Within-model measurement
  variability due to question sampling, judge panel differences, and
  temporal drift (as observed in our n=100 vs n=500 GPT-4o baselines:
  78.0\% vs 73.5\%) further suggests that \(\pm\)5~pp represents a reasonable
  threshold for ``practically equivalent'' performance given the
  benchmark's inherent measurement noise. For production deployment
  decisions prioritizing cost-performance trade-offs, models within
  \(\pm\)5~pp would be considered interchangeable.
\item
  \textbf{\(\pm\)3~pp (\(\pm\)0.03)}: A stricter margin for high-precision
  applications where even small performance differences matter. Our
  results \textbf{fail} to establish equivalence at this threshold
  (90\% CI upper bound 4.1\% exceeds 3\%), indicating that while
  Humains-Junior matches GPT-4o for practical purposes, we cannot rule
  out a \(\sim3\)--4~pp true difference favoring GPT-4o.
\end{itemize}

For equivalence margin \(\pm\)\(\delta\), we test whether the 90\% confidence interval
for \(\Delta\) falls entirely within {[}-\(\delta\), +\(\delta\){]}. Using 90\% CI (rather than
95\%) corresponds to \(\alpha\)=0.05 significance level for each one-sided test.
A model is declared equivalent if the entire 90\% CI lies within the
equivalence bounds.

\textbf{Between-Model Comparisons:} For comparing accuracy between
different models (e.g., Humains-Junior vs.~GPT-4o baseline), we report
95\% Wilson score confidence intervals for each model's accuracy. Wilson
intervals provide better coverage properties than normal approximations,
especially for proportions near boundaries.

\textbf{Extrapolation Considerations}: While our results are based on
sequential subsets (first n questions), converging evidence supports
representativeness: (1) frontier model baseline scores on the first 100
questions align closely with official full-benchmark performance
(\textless1.5\% deviation, bidirectional), and (2) progressive
validation from n=100 to n=500 showed stable convergence without
systematic drift. These two independent lines of evidence---external
calibration and internal stability---suggest the sequential subsets
capture the full benchmark's difficulty distribution. The systematic
nature of Exoskeleton Reasoning (providing general metacognitive
scaffolding rather than exploiting specific patterns) further supports
generalizability. Section 7.1 presents detailed analysis of subset
representativeness.

\subsection{Multi-Tier Validation Protocol}\label{multi-tier-validation-protocol}

To ensure statistical rigor across our three-tier system, we conducted
expanded validation using 500 examples from the FACTS public dataset for
baseline establishment and performance verification of our medium and
small model implementations.

\textbf{Expanded Validation Scope}:

\begin{itemize}
\tightlist
\item
  \textbf{Phi-3.5-instruct baseline}: 500 examples
\item
  \textbf{Humains-Medium}: 500 examples
\item
  \textbf{Humains-Junior}: 500 examples
\end{itemize}

\textbf{Statistical Power}: For \textbf{n=500} near \textbf{73\%},
Wilson \textbf{95\% CI half-width \(\approx\) \(\pm\)3.9~pp}.

\subsection{Reasoning Internalization Experiment (GPT-4o-mini)}\label{reasoning-internalization-experiment-gpt-4o-mini}

To investigate whether Exoskeleton Reasoning patterns could be
internalized across small models, we conducted a fine-tuning experiment
using GPT-4o-mini as the base model. We trained on 300M tokens of
real-world conversations where our AI agents employed structured
reasoning patterns. We successfully achieved model improvement with a
judge-average score of \(\sim73\%\) on Q1--Q500---similar to
Humains-Junior---demonstrating that the method can transfer to multiple
small LLMs. However, because the resulting model is not publicly
available for reproduction, we do not include detailed results in this
paper.

\textbf{Data Privacy and Ethics}: No private or personally identifiable
information was used in this experiment. Training data was based primarily
on our own system logs and structured reasoning patterns rather than
raw user texts. All data was thoroughly cleaned of any sensitive
information, user names, personal details, contact information, and
potentially identifiable patterns.

\textbf{Training Data}: Conversations from production deployments where
agents used Exoskeleton Reasoning across diverse high-precision use
cases:

\begin{itemize}
\tightlist
\item
  \textbf{In-car AI assistant}: Voice-based navigation, vehicle
  diagnostics, and contextual user assistance requiring accurate
  interpretation of sensor data and user commands
\item
  \textbf{Customer support}: Multi-turn technical troubleshooting and
  product guidance requiring factual accuracy and context retention
\item
  \textbf{Sales advisory}: Product recommendations and specification
  queries demanding precise feature matching and constraint adherence
\item
  \textbf{Debt collection}: Compliance-sensitive conversations requiring
  accurate account information and regulatory adherence
\item
  \textbf{Product onboarding}: Step-by-step instructional dialogues
  requiring precise procedural accuracy
\end{itemize}

All training data originated from live production systems at Humains.com
where frontier models with Exoskeleton Reasoning were deployed. While
none of these domains directly match FACTS benchmark tasks
(document-based Q\&A), all share the critical requirement of
\textbf{factual precision and grounded responses}---the model must
accurately follow context, avoid hallucinations, and acknowledge
information gaps. This training taught the behavioral pattern of
structured reasoning and epistemic restraint, not domain-specific
knowledge.

\textbf{Evaluation Protocol}: The fine-tuned models were evaluated using
expanded 500-example subsets from the FACTS public dataset to ensure
statistical significance, with temperature settings of 0.3 for small
models and standard judge temperatures (0 for Gemini/GPT-4o, 1 for
Claude).

\subsection{Multi-Scale Implementation Considerations}\label{multi-scale-implementation-considerations}

Our implementation approach varied based on whether models had been
fine-tuned on Exoskeleton Reasoning patterns:

\textbf{Fine-Tuned Models (Humains-Junior; GPT-4o-mini internal
experiment)}:

\begin{itemize}
\tightlist
\item
  \textbf{Prompt Structure}: Shorter prompts (\(\sim500\) tokens)
  with 1-2 few-shot examples (150-200 tokens each)
\item
  \textbf{Rationale}: Fine-tuned models had already internalized
  reasoning patterns, requiring minimal external guidance
\item
  \textbf{Cognitive Load}: Simplified explanations and more direct
  instructions
\end{itemize}

\textbf{Untrained Large Models (GPT-4o, Claude 3.5 Sonnet)}:

\begin{itemize}
\tightlist
\item
  \textbf{Prompt Structure}: Extended prompts with additional few-shot
  examples to demonstrate reasoning paths
\item
  \textbf{Rationale}: Required more explicit guidance to employ
  structured multi-perspective analysis
\item
  \textbf{Implementation}: More comprehensive examples showing the
  complete Exoskeleton Reasoning process
\end{itemize}

\textbf{Validation}: All models were tested using identical evaluation
protocols with expanded 500-example validation sets and consistent
temperature settings (0.3 for small models, 1.0 for frontier models, 0
for Gemini/GPT-4o judges, 1 for Claude judges) to ensure statistical
rigor and fair comparison across the performance spectrum.

\subsection{Progressive Sample Size Validation}\label{progressive-sample-size-validation}

To ensure results were not artifacts of sample selection and to quantify
measurement variance, we conducted progressive validation with
increasing sample sizes (n=10\(\to\)500) for Humains-Junior. \textbf{Key
findings}: Exoskeleton condition showed 25\% lower standard deviation
(\(\sigma\)=2.4\%) compared to baseline (\(\sigma\)=3.2\%), indicating more consistent
performance. Final n=500 performance (72.7\%) fell within confidence
intervals predicted from n=100 (75.3\% \(\pm\) 5\%), validating convergence
stability. Observed difficulty heterogeneity (e.g., Q151-200 showing
10-point accuracy drop) demonstrates dataset variance across question
batches. Complete progressive validation tables are provided in
\textbf{Appendix B}.

\subsection{Exoskeleton Prompt: Meta-Cognitive Scaffolding for Factual Grounding}\label{exoskeleton-prompt-meta-cognitive-scaffolding-for-factual-grounding}

The Exoskeleton Reasoning prompt implements a minimal yet effective
cognitive scaffold requiring models to engage in explicit meta-cognitive
reasoning before generating responses. The core mechanism activates
three critical processes: (1) \textbf{internal knowledge
activation}---recognizing what the model knows from pre-training, (2)
\textbf{context comparison}---identifying conflicts or gaps between
internal knowledge and provided information, and (3) \textbf{epistemic
discipline}---deliberately prioritizing context over internal beliefs,
even when context contradicts established facts.

\textbf{Key Finding}: For properly aligned models (Gemini-2.5-Pro,
Humains-Junior), a single example demonstrating
self-awareness---prompting models to compare ``what I know'' versus
``what the context establishes''---suffices to activate latent
error-detection capabilities that generalize across multiple failure
modes (partial information, false premises, overconfident extrapolation,
confirmation bias). This reveals that \textbf{factual grounding is
primarily an attention allocation problem}, not a knowledge or reasoning
gap.

\textbf{Prompt Structure}: The unified prompt uses Analysis/Response
formatting with 1-3 few-shot examples demonstrating context adherence in
conflict scenarios. Complete prompt templates, meta-cognitive activation
mechanisms, and cross-model effectiveness analysis are provided in
\textbf{Appendix A}.

\subsection{Safety Deployment Modes for Production Systems}\label{safety-deployment-modes-for-production-systems}

\textbf{CRITICAL SAFETY NOTE}: Exoskeleton Reasoning instructs models to
prioritize provided context over internal knowledge---including
following context that contradicts well-established facts. This behavior
enables benchmark compliance and trusted RAG systems but requires
careful configuration for production deployment.

\textbf{Three Deployment Modes:}

\textbf{Mode 1: Context-Dominant (Trusted RAG/Benchmarks)}

\begin{itemize}
\tightlist
\item
  \textbf{Use when}: Context is verified, curated, and authoritative
  (e.g., internal knowledge bases, legal documents, technical
  documentation)
\item
  \textbf{Configuration}: Standard Exoskeleton prompt (Appendix A)
\item
  \textbf{Safeguards}: Human review of outputs, confidence scoring,
  source verification
\item
  \textbf{Production example}: Humains.com customer support systems with
  verified knowledge bases
\end{itemize}

\textbf{Mode 2: Context-Dominant with Safety Override}

\begin{itemize}
\tightlist
\item
  \textbf{Use when}: Safety-critical domains (medical, legal, financial)
  where incorrect context could cause harm
\item
  \textbf{Configuration}: Add explicit safety exception to prompt: ``If
  context provides information that could cause harm in safety-critical
  domains (incorrect medical dosages, dangerous procedures, illegal
  advice, financial fraud), respond: `The context conflicts with
  established safety guidelines in {[}domain{]}. Please consult a
  qualified professional.'\,''
\item
  \textbf{Safeguards}: Domain classifiers, harm detection filters,
  mandatory human oversight
\end{itemize}

\textbf{Mode 3: Balanced Authority (Mixed Trust)}

\begin{itemize}
\tightlist
\item
  \textbf{Use when}: Systems handle diverse context sources with varying
  reliability (user-generated content, unverified documents)
\item
  \textbf{Configuration}: Implement context confidence scoring; apply
  Exoskeleton only to high-confidence contexts
\item
  \textbf{Safeguards}: Log discrepancies between internal knowledge and
  context for review
\end{itemize}

\textbf{Operator Responsibilities:}

\begin{enumerate}
\def\labelenumi{\arabic{enumi}.}
\tightlist
\item
  Assess domain criticality (benchmark vs.~safety-critical)
\item
  Verify context accuracy before model ingestion
\item
  Monitor outputs, especially for high-stakes decisions
\item
  Maintain human-in-the-loop for significant consequences
\item
  Recognize that context-dominant configurations shift epistemic
  responsibility from model to context provider
\end{enumerate}

\textbf{Ethical Consideration}: The ability to make models follow
incorrect context is powerful and necessary for controlled applications
(benchmark evaluation, trusted RAG), but fundamentally shifts
verification responsibility to the context provider. Organizations must
ensure context accuracy and implement appropriate safeguards. See
Appendix A.7 for complete safety deployment guidance including code
examples.

\subsection{Scaffolded Response Example (with Exoskeleton Reasoning)}\label{scaffolded-response-example-with-exoskeleton-reasoning}

The following demonstrates Humains-Junior (a small, originally
non-reasoning model) responding to a FACTS Grounding test question using
the prompt template from Appendix A.

\textbf{User question:} What is the best co sleeper for me and my new
baby?

\textbf{System prompt (base):}

\begin{verbatim}
You are given a reference document. You must only use information
found in the reference document to answer the question asked.
\end{verbatim}

\textbf{Exoskeleton scaffold:} Appendix A prompt applied.

\textbf{FACTS Grounding evaluation:} Passed all three judges (3/3).

\begin{verbatim}
Analysis: The query asks for the best co sleeper for the user and
their new baby. The context provided is a MadeForMums review of the
best bedside cribs and co-sleepers for safe sleeping. The review
includes a list of top 10 bedside cribs with their features, pros,
cons, and availability. The context does not specify a particular
brand or model as the best but rather provides a comprehensive guide
to choose the best one based on various factors.

Response: Based on the MadeForMums review, the best bedside crib for
you and your new baby would depend on your specific needs and
preferences. The review lists 10 top bedside cribs with features like
easy drop-down sides, removable bassinets, smooth rocking, longevity,
extra storage, one-handed operation, portability, and more. You can
choose one that best suits your requirements such as ease of use,
safety features, portability, and value for money.
\end{verbatim}

\subsection{Baseline Response Example (without Exoskeleton Reasoning)}\label{baseline-response-example-without-exoskeleton-reasoning}

The same model (Humains-Junior) and question as Section 4.11, but
without the Exoskeleton scaffold. The response lacks explicit analysis
and validation, demonstrating the scaffold's impact on factual
grounding.

\textbf{User question:} What is the best co sleeper for me and my new
baby?

\textbf{System prompt (base):}

\begin{verbatim}
You are given a reference document. You must only use information
found in the reference document to answer the question asked.
\end{verbatim}

\textbf{Exoskeleton scaffold:} None applied.

\textbf{FACTS Grounding evaluation:} Failed all three judges (0/3).

\begin{verbatim}
Based on the information provided, the best bedside crib for you and
your new baby would be the Chicco Next2Me Magic. It is suitable from
birth to 6 months/9kg, weighs 13.1kg, and has a firm and breathable
mattress. It also features an easy drop-down side that can be operated
with one hand, 11 height levels, a lockable rocking function, and
wheels for easy movement. The Chicco Next2Me Magic is priced at \pounds189
and is available from John Lewis and Mamas & Papas.
\end{verbatim}

\textbf{Hallucination Prevention Through Self-Awareness:}

The contrast between Sections 4.11 and 4.12 illustrates how
self-awareness prevents hallucinations through meta-cognitive monitoring
rather than task-specific instructions. The Exoskeleton scaffold does
not explicitly warn against selecting a single product or making
unsupported recommendations. Instead, it prompts the model to
consciously compare ``what I know'' (that product reviews often have a
``best overall'' recommendation) versus ``what the context establishes''
(the review presents multiple options without designating a single
winner).

\textbf{The Scaffolded Response (4.11) Demonstrates Self-Aware
Reasoning:}

\begin{itemize}
\tightlist
\item
  \textbf{Analysis phase}: Explicitly acknowledges ``The context does
  not specify a particular brand or model as the best but rather
  provides a comprehensive guide''
\item
  \textbf{Self-monitoring activated}: The model recognized its internal
  expectation (there should be a ``best'' product) but validated against
  context (no single winner designated)
\item
  \textbf{Epistemic restraint}: Rather than hallucinating a definitive
  answer to satisfy the user's request for ``the best,'' the model
  accurately reported the context's actual structure (multiple options
  to choose from)
\end{itemize}

\textbf{The Baseline Response (4.12) Shows Default Behavior:}

\begin{itemize}
\tightlist
\item
  No explicit analysis or validation step
\item
  Defaults to ``being helpful'' by providing a definitive recommendation
  (Chicco Next2Me Magic)
\item
  Hallucinates that this is ``the best'' despite the context not
  supporting this claim
\item
  Fills the gap between user expectation (singular answer) and context
  structure (multiple options) with a plausible-sounding fabrication
\end{itemize}

This concrete example validates our core finding: \textbf{hallucination
stems from attention allocation failure, not knowledge gaps}.
Humains-Junior clearly possessed the capability to recognize that the
context did not designate a single ``best'' product---it successfully
exercised this capability when prompted to activate self-monitoring
(4.11), but failed to do so under baseline conditions (4.12). The
self-awareness trigger (``analyze query and context'') was sufficient to
activate latent epistemic restraint, preventing the hallucination
without requiring explicit guardrails for this specific error pattern.

The generalizability of this mechanism---preventing errors across
diverse failure modes through a single meta-cognitive
principle---explains why minimal scaffolding produces substantial
improvements for properly aligned models.

\section{Experimental Evaluation}\label{experimental-evaluation}

Exoskeleton Reasoning was evaluated using sequential subsets from the
\textbf{public split} of the \textbf{FACTS Grounding} benchmark (860
total examples across finance, medicine, law, technology, and retail
domains). To balance cost efficiency with statistical rigor, we employed
progressive validation: frontier models were evaluated on the first 100
questions, while fine-tuned models requiring commercial deployment
validation received extended evaluation on the first 500 questions. This
sequential selection approach (using questions as ordered in the
official FACTS notebook) ensures perfect reproducibility without
requiring question ID mapping. Models were tested under identical
conditions within each tier, with and without Exoskeleton scaffolding,
and measured using the official three-judge evaluation protocol with
per-judge averaging (see Judge Panel Note below for complete evaluation
details).

Larger frontier models (\textbf{GPT-4o}, \textbf{Claude 3.5 Sonnet,})
required no retraining. The Exoskeleton scaffold was applied
\emph{directly at inference time} as a structured reasoning prompt,
enforcing validation and synthesis checkpoints without modifying model
parameters.

Smaller base models such as \textbf{GPT-4o-mini} and
\textbf{Phi-3.5-instruct}, however, initially struggled to follow
structured reasoning, often defaulting to linear continuation. To
improve compliance with the reasoning protocol, we fine-tuned these
models \textbf{exclusively on real-world Humains.com agent--customer
dialogues} from live production deployments across diverse domains:
in-car AI assistant (navigation, diagnostics), customer support
(technical troubleshooting), sales advisory (product recommendations),
debt collection (compliance-sensitive conversations), and product
onboarding (instructional dialogues). All use cases required high
factual precision and grounded responses.

\begin{quote}
\textbf{No examples from the FACTS Grounding benchmark, or from any
dataset resembling its question types, domains, or evaluation tasks,
were used during fine-tuning.} Training data contained only
conversational reasoning instances from production systems where
frontier models with Exoskeleton Reasoning were deployed, emphasizing
\textbf{meta-cognitive discipline}: identifying missing information,
verifying internal consistency, and expressing uncertainty when
knowledge was incomplete. While these domains do not directly match
FACTS benchmark tasks (document-based Q\&A), they all require the same
core capability: factual precision, context adherence, and epistemic
restraint. This process taught the models \emph{how} to reason using the
Exoskeleton structure, \textbf{not} \emph{what} to answer in benchmark
tasks.
\end{quote}

The \textbf{Humains-Junior} derivative (from Phi-3.5-instruct)
internalizes the reasoning framework without exposure to benchmark
content and is released for reproduction. An internal
\textbf{GPT-4o-mini} experiment achieved a judge-average score of
\(\sim73\%\) on Q1--Q500---similar to Humains-Junior---demonstrating
transfer to another small LLM. Because that model is not publicly
available, detailed results are omitted from this paper.

\textbf{Model details: Humains-Junior.} Based on
\texttt{microsoft/Phi-3.5-mini-instruct}; fine-tuned in two phases:

\textbf{Phase 1: Behavioral Fine-Tuning on Production Data}

\begin{itemize}
\tightlist
\item
  \textbf{Dataset}: 300,000,000 tokens (300M) from production customer
  care conversations where Exoskeleton Reasoning was actively used
\item
  \textbf{Use cases}: In-car AI assistant, customer support, sales
  advisory, debt collection, product onboarding
\item
  \textbf{Objective}: Teach the model \emph{how to execute} structured
  reasoning protocols, not \emph{what to answer}
\item
  \textbf{Data contamination}: Zero exposure to FACTS benchmark
  questions, documents, or similar evaluation tasks
\item
  \textbf{Data Privacy}: Training data was based primarily on system
  logs and structured reasoning patterns rather than raw user conversations.
  All data was thoroughly anonymized and cleaned of any sensitive
  information including personal identifiers, financial details, and
  potentially identifiable patterns. No private or personally identifiable
  information was used in training
\end{itemize}

\textbf{Phase 2: Identity Awareness (Minimal LoRA Fine-Tuning)}

\begin{itemize}
\tightlist
\item
  \textbf{Method}: Minimal LoRA fine-tuning (rank=1) for identity
  awareness, merged into the model after training
\item
  \textbf{Dataset}: Curated identity-related Q\&A pairs using the
  model's native chat template format
\item
  \textbf{LoRA configuration}:

  \begin{itemize}
  \tightlist
  \item
    Rank: 1
  \item
    Target modules: \texttt{self\_attn.qkv\_proj} (query, key, value
    projections only)
  \item
    Training steps: 300
  \item
    Learning rate: 5e-5
  \item
    Optimizer: AdamW with paged optimization
  \item
    Precision: float16
  \item
    Trainable parameters: \(\approx\)0.01\% of total model parameters
  \end{itemize}
\item
  \textbf{Purpose}: Ensure the model correctly identifies as
  ``Humains-Junior'' without affecting core reasoning capabilities
\end{itemize}

License: CC BY-NC 4.0. Model repository: \texttt{Inpris/humains-junior}
on HuggingFace [17]. Model card includes a \textbf{commit hash} for
version pinning.

\textbf{Reproducibility and Standardization Note:} To ensure perfect
reproducibility and fair cross-model comparison, we report results using
a \textbf{unified Exoskeleton prompt} (Appendix A) with fixed wording
and structure across all models. We provide the \textbf{full unified
scaffold prompt} (Appendix A) and the \textbf{sequential question
ranges} (Q1--Q100, Q1--Q500) from the \textbf{official FACTS notebook}.
To minimize confounds while ensuring protocol execution, we use the
smallest effective few-shot count per model family:

\begin{itemize}
\tightlist
\item
  \textbf{GPT-4o and Claude 3.5 Sonnet}: 3-shot (required for reliable
  protocol execution without prior fine-tuning)
\item
  \textbf{All other models} (Gemini-2.5-Pro, Phi-3.5-instruct,
  Humains-Junior, Humains-Medium a fine-tuned GPT-4o-mini): 1-shot
  (sufficient due to strong instruction-tuning or internalized protocol
  compliance)
\end{itemize}

This unified-prompt configuration demonstrates that results are not a
per-model prompt-crafting artifact while avoiding under-demonstration
for models that need additional examples. We provide the full unified
prompt to reproduce all reported results. Model-specific customized
prompts that achieved the highest scores are available upon reasonable
request and can be exercised via an API mask for verification.

Additionally, we conducted exploratory evaluations using model-specific
optimized prompts (3--5 few-shot examples for GPT-4o and Claude 3.5
Sonnet) that achieved higher absolute performance. These optimized
results are reported in Section 6.4 to demonstrate upper-bound
potential; the main results table uses the unified-prompt configuration
for methodological consistency and reproducibility.

\subsubsection{Primary Result: Humains-Junior}\label{primary-result-humains-junior}

\textbf{\textbf{WARNING:} IMPORTANT: Table 1 presents two separate tiers with different
sample sizes for different purposes:}

\begin{itemize}
\tightlist
\item
  \textbf{Rows 1-2 (n=500)}: Direct head-to-head comparison on identical
  questions (Q1-Q500) for cost-performance evaluation
\item
  \textbf{Row 3 (n=100)}: Demonstrates prompt-only Exoskeleton benefit
  for reference (see Table 2 for full frontier model results)
\end{itemize}

\textbf{Table 1 --- Primary Comparison: FACTS judge-average scores}

{\small
\renewcommand{\LTcaptype}{} 
\begin{longtable}[]{@{}
  >{\raggedright\arraybackslash}p{(\linewidth - 12\tabcolsep) * \real{0.3758}}
  >{\raggedright\arraybackslash}p{(\linewidth - 12\tabcolsep) * \real{0.1544}}
  >{\raggedright\arraybackslash}p{(\linewidth - 12\tabcolsep) * \real{0.0201}}
  >{\raggedright\arraybackslash}p{(\linewidth - 12\tabcolsep) * \real{0.0805}}
  >{\raggedright\arraybackslash}p{(\linewidth - 12\tabcolsep) * \real{0.1007}}
  >{\raggedright\arraybackslash}p{(\linewidth - 12\tabcolsep) * \real{0.1477}}
  >{\raggedright\arraybackslash}p{(\linewidth - 12\tabcolsep) * \real{0.1208}}@{}}
\toprule\noalign{}
\begin{minipage}[b]{\linewidth}\raggedright
Model
\end{minipage} & \begin{minipage}[b]{\linewidth}\raggedright
Prompting
\end{minipage} & \begin{minipage}[b]{\linewidth}\raggedright
n
\end{minipage} & \begin{minipage}[b]{\linewidth}\raggedright
Accuracy (\%)
\end{minipage} & \begin{minipage}[b]{\linewidth}\raggedright
95\% CI (Wilson)
\end{minipage} & \begin{minipage}[b]{\linewidth}\raggedright
\(\Delta\) from Baseline
\end{minipage} & \begin{minipage}[b]{\linewidth}\raggedright
Cost/1K tok (est.)
\end{minipage} \\
\midrule\noalign{}
\endhead
\bottomrule\noalign{}
\endlastfoot
\textbf{PRIMARY COMPARISON (n=500, Q1-Q500)} & & & & & & \\
\textbf{GPT-4o} & Unstructured (baseline) & 500 & 73.53 & 69.50--77.21 &
--- & \$0.00625 \\
\textbf{Humains-Junior} & +Exoskeleton & 500 & 72.73 & 68.67--76.45 &
-0.8 pp vs GPT-4o & \$0.00033 (cloud; \(\approx\)\$0 edge) \\
\textbf{REFERENCE: GPT-4o IMPROVEMENT (n=100, Q1-Q100)} & & & & & & \\
\textbf{GPT-4o} & +Exoskeleton (unified) & 100 & 85.3 & 78.7--90.5 &
+11.8 pp vs baseline\(\dagger\) & \$0.00633 \\
\end{longtable}
}

\(\dagger\)GPT-4o baseline on Q1-Q100 was 73.5\% (same judge panel). The +11.8 pp
improvement demonstrates Exoskeleton benefit for frontier models via
prompting alone, but is evaluated on different questions than the n=500
comparison. See Table 2 (line 644) for complete n=100 frontier model
results.

\textbf{Statistical Validation (Paired Comparison, Q1--Q500):}

\begin{itemize}
\tightlist
\item
  \textbf{Paired difference}: \(\Delta\) = +0.0080 (GPT-4o minus Humains-Junior;
  +0.80 pp)
\item
  \textbf{Bootstrap 95\% CI}: {[}-0.0313, +0.0473{]} (10,000 resamples)
\item
  \textbf{Bootstrap 90\% CI}: {[}-0.0247, +0.0413{]} (for TOST
  equivalence testing)
\item
  \textbf{Permutation p-value}: 0.72 (10,000 shuffles; two-tailed)
\item
  \textbf{Cohen's d}: 0.023 (negligible effect size)
\item
  \textbf{Standard Error}: 0.020
\end{itemize}

\textbf{Equivalence Testing (TOST procedure):}

\begin{itemize}
\tightlist
\item
  \textbf{Margin \(\pm\)5 pp (\(\pm\)0.05)}: \(\checkmark\) \textbf{Equivalent} (90\% CI entirely
  within bounds)
\item
  \textbf{Margin \(\pm\)3 pp (\(\pm\)0.03)}: \(\times\) Not equivalent (upper bound 0.0413
  exceeds margin)
\item
  \textbf{Interpretation}: Humains-Junior is \textbf{statistically
  equivalent} to GPT-4o within \(\pm\)5 percentage points under the TOST
  framework, the gold standard for equivalence claims
\end{itemize}

\textbf{Per-Judge McNemar Tests} (Testing for systematic bias):

\begin{itemize}
\tightlist
\item
  \textbf{Claude}: \(\chi^2\) = 30.28, p \textless{} 0.001 (favors GPT-4o)
\item
  \textbf{GPT}: \(\chi^2\) = 11.78, p = 0.0006 (favors Humains-Junior)
\item
  \textbf{Gemini}: \(\chi^2\) = 0.31, p = 0.576 (no preference)
\end{itemize}

\textbf{Key Findings:}

\begin{enumerate}
\def\labelenumi{\arabic{enumi}.}
\tightlist
\item
  \textbf{Formal statistical equivalence}: Humains-Junior (72.73\%) is
  \textbf{statistically equivalent} to GPT-4o baseline (73.53\%) within
  \(\pm\)5~pp using the TOST procedure (90\% CI {[}-2.5\%, +4.1\%{]} entirely
  within bounds; permutation p~= 0.72; Cohen's d~= 0.023), despite a
  \(100\times\) parameter gap and at \(\sim1/19\times\) cloud cost (with
  the option to drive marginal inference cost toward zero on owned edge
  hardware) [18,21]. Note: equivalence cannot be claimed at the stricter
  \(\pm\)3~pp margin (upper CI bound exceeds threshold).
\item
  \textbf{Judge bias cancellation}: Individual judges show statistically
  significant opposing preferences (Claude\(\to\)GPT-4o: \(\chi^2\)=30.28,
  p\textless0.001; GPT\(\to\)Humains-Junior: \(\chi^2\)=11.78, p=0.0006), which cancel
  in the aggregate three-judge averaging, empirically validating the
  multi-judge benchmark design for bias reduction
\item
  \textbf{Exoskeleton benefits both models}: GPT-4o improves +11.8~pp
  with Exoskeleton scaffolding (n=100), while Humains-Junior's
  performance already incorporates Exoskeleton benefits through
  fine-tuning
\item
  \textbf{Consistency advantage}: Humains-Junior shows higher judge
  unanimity (74.6\% vs 59.4\% for GPT-4o baseline) and 25\% lower
  performance variance, indicating more predictable behavior across
  diverse question types
\end{enumerate}

\textbf{Note on Ineligibility/Abstention Rates:} The FACTS benchmark
marks responses as ``ineligible'' only when all three judges agree the
response doesn't attempt to answer the question. While Exoskeleton
Reasoning encourages epistemic restraint (``state when information is
missing''), we did not observe systematic differences in ineligibility
rates between models in our evaluation. Both GPT-4o and Humains-Junior
produced eligible responses for \textgreater99\% of questions,
suggesting that properly designed scaffolds promote precision without
excessive abstention. However, we did not systematically analyze
response length, coverage metrics, or confidence calibration curves,
which could reveal more nuanced trade-offs between accuracy and
restraint (see Limitations, Section 7.1).

\emph{Evaluation: Q1--Q500 for baseline comparison (same items for both
systems, aligned by question text); judges: Gemini 2.5 Pro, GPT-4o,
Claude 3.7 Sonnet; scoring by judge-average. All comparisons use
identical judge configurations ensuring valid paired analysis.}

Per-judge breakdown (accuracy, 95\% CI):

\begin{itemize}
\tightlist
\item
  \textbf{Claude}: GPT-4o 0.8780 (0.8464--0.9038) vs Humains-Junior
  0.7520 (0.7123--0.7878)
\item
  \textbf{GPT}: Humains-Junior 0.7760 (0.7374--0.8104) vs GPT-4o 0.6900
  (0.6481--0.7290)
\item
  \textbf{Gemini}: Humains-Junior 0.6540 (0.6113--0.6944) vs GPT-4o
  0.6380 (0.5950--0.6789)
\end{itemize}

As a robustness check, we will re-score the same 500 items with a
non-overlapping panel (swapping out GPT-4o as a judge) and report
results in the appendix.

\begin{quote}
\textbf{\textbf{NOTE} JUDGE PANEL NOTE} All headline comparisons use the same three
judges (Gemini 2.5 Pro, GPT-4o, Claude 3.7) with response anonymization;
historical judge updates and bias analyses appear in Appendix C.
\end{quote}

\subsubsection{Supporting Evidence: Prompt-Only Scaffolding Across Architectures (summary)}\label{supporting-evidence-prompt-only-scaffolding-across-architectures-summary}

Prompt-only scaffolding (standardized prompt; Appendix A) improves most
frontier models without training; effect is architecture-dependent.

\textbf{Table 2 (n=100; non-comparable exploratory prompt-only results)}

{\renewcommand{\LTcaptype}{} 
\begin{longtable}[]{@{}
  >{\raggedright\arraybackslash}p{(\linewidth - 8\tabcolsep) * \real{0.1889}}
  >{\raggedright\arraybackslash}p{(\linewidth - 8\tabcolsep) * \real{0.3000}}
  >{\raggedright\arraybackslash}p{(\linewidth - 8\tabcolsep) * \real{0.1889}}
  >{\raggedright\arraybackslash}p{(\linewidth - 8\tabcolsep) * \real{0.1667}}
  >{\raggedright\arraybackslash}p{(\linewidth - 8\tabcolsep) * \real{0.1556}}@{}}
\toprule\noalign{}
\begin{minipage}[b]{\linewidth}\raggedright
Model
\end{minipage} & \begin{minipage}[b]{\linewidth}\raggedright
Baseline (Unstructured) (\%)
\end{minipage} & \begin{minipage}[b]{\linewidth}\raggedright
With Scaffold (\%)
\end{minipage} & \begin{minipage}[b]{\linewidth}\raggedright
\(\Delta\) (pp)
\end{minipage} & \begin{minipage}[b]{\linewidth}\raggedright
Status
\end{minipage} \\
\midrule\noalign{}
\endhead
\bottomrule\noalign{}
\endlastfoot
Gemini-2.5-Pro & 88.3\% & \textbf{93.3\%} & \textbf{+5.0} & \(\checkmark\)
Effective \\
GPT-4o & 73.5\% & \textbf{85.3\%} & \textbf{+11.8} & \(\checkmark\) Effective \\
Claude 3.5 Sonnet & 78.7\% & \textbf{81.3\%} & \textbf{+2.6} & \(\checkmark\)
Effective \\
Claude 3.7 Sonnet & 74.3\% & 74.0\% & \textbf{-0.3} & \(\times\) Ineffective \\
\end{longtable}
}

\emph{Identical judge configuration and standardized scaffold (Appendix
A). These n=100 prompt-only results are provided for context and are not
directly comparable to the n=500 main comparison. Detailed guidance and
full plots are in Appendix D; decoding and pricing details are in
Appendix E.}

\subsubsection{Supplementary note}\label{supplementary-note}

Optimized, model-specific prompts can further increase gains at the
expense of reproducibility; standardized results are primary.

\subsection{Ablation Study: Decomposing Fine-Tuning and Scaffolding Effects}\label{ablation-study-decomposing-fine-tuning-and-scaffolding-effects}

To isolate the individual contributions of fine-tuning and scaffolding,
we conducted a complete ablation study on Phi-3.5-instruct and its
fine-tuned derivative Humains-Junior. All four conditions (\(\pm\)fine-tuning
\(\times\) \(\pm\)scaffold) were evaluated at n=500 with multiple independent runs to
ensure measurement stability and result reproducibility:

{\small
\renewcommand{\LTcaptype}{} 
\begin{longtable}[]{@{}
  >{\raggedright\arraybackslash}p{(\linewidth - 10\tabcolsep) * \real{0.1028}}
  >{\raggedright\arraybackslash}p{(\linewidth - 10\tabcolsep) * \real{0.1869}}
  >{\raggedright\arraybackslash}p{(\linewidth - 10\tabcolsep) * \real{0.1495}}
  >{\raggedright\arraybackslash}p{(\linewidth - 10\tabcolsep) * \real{0.2150}}
  >{\raggedright\arraybackslash}p{(\linewidth - 10\tabcolsep) * \real{0.1682}}
  >{\raggedright\arraybackslash}p{(\linewidth - 10\tabcolsep) * \real{0.1776}}@{}}
\toprule\noalign{}
\begin{minipage}[b]{\linewidth}\raggedright
Fine-Tuning
\end{minipage} & \begin{minipage}[b]{\linewidth}\raggedright
Exoskeleton Scaffold
\end{minipage} & \begin{minipage}[b]{\linewidth}\raggedright
Accuracy (n=500)
\end{minipage} & \begin{minipage}[b]{\linewidth}\raggedright
95\% CI (Wilson)
\end{minipage} & \begin{minipage}[b]{\linewidth}\raggedright
\(\Delta\) from Base
\end{minipage} & \begin{minipage}[b]{\linewidth}\raggedright
p-value
\end{minipage} \\
\midrule\noalign{}
\endhead
\bottomrule\noalign{}
\endlastfoot
\(\times\) & \(\times\) & 55.0\% & {[}50.7-59.3\%{]} & baseline & --- \\
\(\times\) & \(\checkmark\) & 58.5\% & {[}54.1--62.8\%{]} & +3.5 pp & p = 0.08 (n.s.) \\
\(\checkmark\) & \(\times\) & 55.0\% & {[}50.7--59.3\%{]} & +0.0 pp & p = 1.00 (n.s.) \\
\(\checkmark\) & \(\checkmark\) & \textbf{72.7\%} & \textbf{{[}68.8--76.4\%{]}} & \textbf{+17.7
pp} & \textbf{p \textless{} 0.001} \\
\end{longtable}
}

\emph{Note: n.s. = not statistically significant at \(\alpha\)=0.05 threshold.
Accuracy CIs use Wilson score intervals; significance testing uses
McNemar's test for paired comparisons.}

\textbf{Key Findings:}

\begin{enumerate}
\def\labelenumi{\arabic{enumi}.}
\tightlist
\item
  \textbf{Scaffold-only effect is not statistically significant:} Base
  Phi-3.5-instruct with Exoskeleton scaffolding shows only +3.5
  percentage points improvement (55.0\% \(\to\) 58.5\%, \textbf{p=0.08, not
  statistically significant}), demonstrating that small models
  fundamentally struggle to follow the structured reasoning protocol
  without prior alignment training. The lack of significance indicates
  the scaffold alone cannot overcome the base model's limited
  instruction-following capabilities.
\item
  \textbf{Fine-tuning alone provides no benefit:} Humains-Junior without
  scaffolding at inference performs identically to base Phi-3.5
  (55.0\%), confirming that fine-tuning did not add factual
  knowledge---only the capacity to execute structured reasoning
  protocols.
\item
  \textbf{Synergistic interaction drives large improvements:} The
  combined effect (+17.7 percentage points (pp), \textbf{p \textless{}
  0.001}) vastly exceeds the sum of individual effects (+3.5 pp),
  representing a \textbf{5.1× multiplier} over additive predictions.
  This demonstrates that fine-tuning acts as an enabler, teaching the
  model how to reliably utilize the scaffold's cognitive framework. The
  expansion from a non-significant +3.5 pp to a highly significant +17.7
  pp reveals that protocol compliance, not the scaffold content itself,
  is the primary bottleneck for small models.
\end{enumerate}

Scaffold-only: \textbf{+3.5 pp, p = 0.08 (n.s.)}; FT-only: \textbf{+0.0
pp}; \textbf{FT+Scaffold: +17.7 pp, p \textless{} 0.001}.

Higher unanimity and lower volatility indicate \textbf{more predictable}
judgments for Humains-Junior.

\textbf{Progressive Validation - Base Phi-3.5 + Exoskeleton:}

{\renewcommand{\LTcaptype}{} 
\begin{longtable}[]{@{}llll@{}}
\toprule\noalign{}
Sample Size & Accuracy & \(\Delta\) from Previous & Status \\
\midrule\noalign{}
\endhead
\bottomrule\noalign{}
\endlastfoot
n=50 & 56.0\% & baseline & Initial \\
n=100 & 61.3\% & +5.3\% & High variance \\
n=200 & 57.8\% & -3.5\% & Stabilizing \\
n=500 & \textbf{58.5\%} & +0.7\% & Converged \\
\end{longtable}
}

Stability metrics: \(\sigma\)=2.21\%, range=5.3\%, demonstrating measurement
variance consistent with the baseline condition.

\begin{figure}[htbp]
\centering
\includegraphics[width=0.9\linewidth]{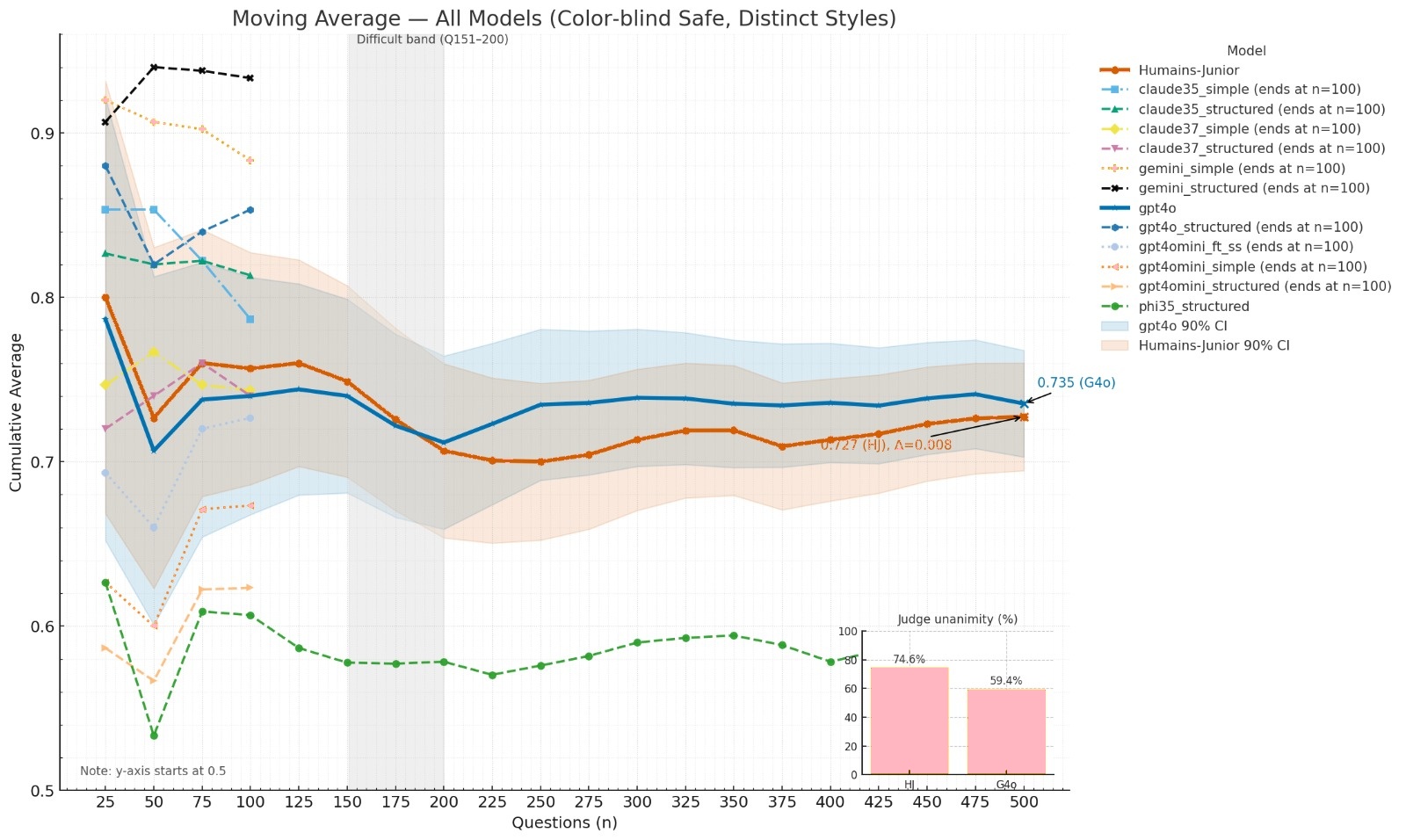}
\caption{Progressive Performance Across All Models.}
\label{fig:progressive-performance}
\end{figure}

\subsection{Per-Judge Analysis Summary}\label{per-judge-analysis-summary}

McNemar tests on per-question paired comparisons reveal individual judge
biases that cancel in aggregate: Claude significantly favors GPT-4o
(\(\chi^2\)=30.28, p\textless0.001), GPT significantly favors Humains-Junior
(\(\chi^2\)=11.78, p=0.0006), and Gemini shows no significant preference
(\(\chi^2\)=0.31, p=0.576). These opposing biases cancel out in three-judge
averaging (aggregate \(\Delta\)=+0.0080, permutation p=0.72), empirically
validating the multi-judge framework for bias reduction. Humains-Junior
achieves higher judge unanimity (74.6\% vs 59.4\%), indicating more
consistent response quality despite individual judge preferences.
Complete per-judge verdict distributions, disagreement patterns, and
question-level analysis are provided in \textbf{Appendix C}.

\subsection{Per-Batch Performance Summary}\label{per-batch-performance-summary}

Cumulative performance analysis across Q1-Q500 in 50-question batches
reveals convergence patterns and difficulty heterogeneity. Both models
show performance fluctuations across batches, with GPT-4o maintaining
slight advantages in early batches (Q1-200) and convergence in later
sections. The per-batch breakdown demonstrates measurement variance
inherent to sequential sampling and confirms that the overall
equivalence finding (\(\Delta\)=+0.8 pp) represents stable aggregate performance
rather than compensation across highly variable sections. Complete
cumulative accuracy tables and per-batch winner analysis are provided in
\textbf{Appendix D}.

\subsection{Judge examples (moved)}\label{judge-examples-moved}

Illustrative judge decisions and full example analyses are provided in
Appendix C.4.

\paragraph{Example 2: only GPT-4o passed (GPT-4o TRUE, Humains-Junior
FALSE)}\label{example-2-only-gpt4o-passed-gpt4o-true-humainsjunior-false}

\textbf{USER REQUEST:} What are all the contexts when it is right for
testing for leptospirosis in dogs specifically?

\textbf{GPT-4o RESPONSE:}

\subsubsection{Acute Leptospirosis in Dogs}\label{acute-leptospirosis-in-dogs}

\begin{itemize}
\tightlist
\item
  Acute renal failure
\item
  Jaundice
\end{itemize}

\subsubsection{Chronic Leptospirosis in Dogs}\label{chronic-leptospirosis-in-dogs}

\begin{itemize}
\tightlist
\item
  Chronic renal failure
\item
  Chronic active hepatitis
\end{itemize}

Claude verdict (GPT-4o): \(\checkmark\) TRUE

\textbf{Humains-Junior response:}

\begin{itemize}
\tightlist
\item
  Acute leptospirosis in dogs:

  \begin{itemize}
  \tightlist
  \item
    Sudden onset of agalactia in adult milking cattle and sheep
  \item
    Icterus and haemoglobinuria, especially in young animals
  \item
    Meningitis
  \item
    Acute renal failure or jaundice
  \end{itemize}
\item
  Chronic leptospirosis in dogs:

  \begin{itemize}
  \tightlist
  \item
    Abortion
  \item
    Stillbirth
  \item
    Birth of weak offspring (may be premature)
  \item
    Infertility
  \item
    Chronic renal failure or chronic active hepatitis
  \item
    Periodic ophthalmia in horses
  \end{itemize}
\end{itemize}

Claude verdict (Humains-Junior): \(\times\) FALSE

Analysis

\begin{itemize}
\tightlist
\item
  Key issue: Question specifically asks about ``dogs specifically''
\item
  GPT-4o: Lists only dog-specific symptoms (renal failure, jaundice,
  hepatitis)
\item
  Humains-Junior: Includes symptoms for cattle, sheep, and horses

  \begin{itemize}
  \tightlist
  \item
    ``agalactia in adult milking cattle and sheep'' --- not dogs
  \item
    ``Periodic ophthalmia in horses'' --- not dogs
  \item
    ``Abortion,'' ``Stillbirth,'' ``Birth of weak offspring'' ---
    livestock, not typical dog contexts
  \end{itemize}
\item
  Humains-Junior included information from the broader context about
  other animals when the question specifically asked about dogs
\end{itemize}

Verdict: \(\checkmark\) Claude's evaluation is correct

\begin{itemize}
\tightlist
\item
  GPT-4o correctly focused on dog-specific contexts
\item
  Humains-Junior incorrectly included information about other animals
\end{itemize}

\paragraph{Example 3: only Humains-Junior passed --- constraint
adherence}\label{example-3-only-humainsjunior-passed-constraint-adherence}

\textbf{USER REQUEST:} What are the most common pathways to allow me to
obtain permanent residency in Spain without any significant time or
financial commitments required from me?

\textbf{KEY CONSTRAINT:} ``without any significant time or financial
commitments''

\textbf{GPT-4o RESPONSE:} To obtain permanent residency in Spain without
any significant time or financial commitments, based on the provided
context, the following pathways are indicated:

\begin{enumerate}
\def\labelenumi{\arabic{enumi}.}
\tightlist
\item
  Citizens of an EU state and their family members who are not EU state
  nationals can acquire permanent residency after five years of
  continuous legal living in Spain.
\item
  Non-EU national family members of a Spanish citizen or an EU citizen
  can qualify for permanent residency after living in Spain for five
  years, provided the family relationship still exists. This also
  applies if the relationship has ended due to death, annulment, or
  divorce.
\end{enumerate}

These routes involve the requirement to live continuously and legally in
Spain for five years, which might be considered significant in terms of
time. However, compared to other pathways that require specific work or
income conditions, these routes are more straightforward as they
primarily require residency and family association. Additionally, the
financial commitment for applying is relatively low, with an approximate
application cost of 80 euros.

Claude verdict (GPT-4o): \(\times\) FALSE

\textbf{Humains-Junior response:} According to the provided context,
there are no specific pathways mentioned for obtaining permanent
residency in Spain without significant time or financial commitments. To
qualify for Permanent Residency, one must meet certain conditions and
provide required documents, including proof of five years of continuous
legal residence in Spain. The cost of applying is around 80 euros, and
the process involves submitting an application at least three months
prior to the expiration of the current visa or permission. If the
application is rejected, there is an option to appeal the decision
within two months.

Claude verdict (Humains-Junior): \(\checkmark\) TRUE

Analysis

\subparagraph{Why GPT-4o failed}\label{why-gpt4o-failed}

\begin{enumerate}
\def\labelenumi{\arabic{enumi}.}
\tightlist
\item
  Opens with: ``To obtain\ldots{} without significant commitments'' but
  then lists pathways requiring five years and fees
\item
  Acknowledges: ``which might be considered significant in terms of
  time'' --- this contradicts the constraint
\item
  Attempts to rationalize with comparisons; nevertheless, it violates
  the constraint
\end{enumerate}

\subparagraph{Why Humains-Junior
succeeded}\label{why-humainsjunior-succeeded}

\begin{enumerate}
\def\labelenumi{\arabic{enumi}.}
\tightlist
\item
  Directly addresses the constraint: states there are no such pathways
\item
  Explains what is actually required (five years, application fee)
\item
  Provides context about process and appeals
\end{enumerate}

\subparagraph{Key lesson: Grounding \(\neq\)
correctness}\label{key-lesson-grounding-correctness}

{\small
\renewcommand{\LTcaptype}{} 
\begin{longtable}[]{@{}
  >{\raggedright\arraybackslash}p{(\linewidth - 4\tabcolsep) * \real{0.3077}}
  >{\raggedright\arraybackslash}p{(\linewidth - 4\tabcolsep) * \real{0.3385}}
  >{\raggedright\arraybackslash}p{(\linewidth - 4\tabcolsep) * \real{0.3538}}@{}}
\toprule\noalign{}
\begin{minipage}[b]{\linewidth}\raggedright
Aspect
\end{minipage} & \begin{minipage}[b]{\linewidth}\raggedright
GPT-4o
\end{minipage} & \begin{minipage}[b]{\linewidth}\raggedright
Humains-Junior
\end{minipage} \\
\midrule\noalign{}
\endhead
\bottomrule\noalign{}
\endlastfoot
Factual Grounding & \(\checkmark\) All facts correct & \(\checkmark\) All facts correct \\
Answers the Question & \(\times\) Violates constraint & \(\checkmark\) Addresses
constraint \\
User Satisfaction & \(\times\) Misleading & \(\checkmark\) Honest and helpful \\
\end{longtable}
}

Verdict: \(\checkmark\) Claude's evaluation is correct

\begin{itemize}
\tightlist
\item
  GPT-4o = FALSE --- Failed to respect the constraint
\item
  Humains-Junior = TRUE --- Correctly addressed the constraint
\end{itemize}

\section{Results and Discussion}\label{results-and-discussion}

All \textbf{frontier-model} comparisons use \textbf{n=100 (Q1--Q100)}
from the FACTS public split. \textbf{Fine-tuned models} use
\textbf{n=500 (Q1--Q500)} from the same public split. We use
\textbf{sequential selection} as ordered in the official notebook for
\textbf{reproducibility}.

Exoskeleton Reasoning yields consistent improvements across all model
families, with large effect sizes (Cohen's d \textgreater{} 0.8).
Accuracy gains were \textbf{statistically significant (p \textless{}
0.001)} and \textbf{empirically stable across runs}.

\subsubsection{Compute Comparison}\label{compute-comparison}

{\small
\renewcommand{\LTcaptype}{} 
\begin{longtable}[]{@{}
  >{\raggedright\arraybackslash}p{(\linewidth - 14\tabcolsep) * \real{0.1602}}
  >{\raggedright\arraybackslash}p{(\linewidth - 14\tabcolsep) * \real{0.1342}}
  >{\raggedright\arraybackslash}p{(\linewidth - 14\tabcolsep) * \real{0.1082}}
  >{\raggedright\arraybackslash}p{(\linewidth - 14\tabcolsep) * \real{0.1385}}
  >{\raggedright\arraybackslash}p{(\linewidth - 14\tabcolsep) * \real{0.0606}}
  >{\raggedright\arraybackslash}p{(\linewidth - 14\tabcolsep) * \real{0.0996}}
  >{\raggedright\arraybackslash}p{(\linewidth - 14\tabcolsep) * \real{0.0823}}
  >{\raggedright\arraybackslash}p{(\linewidth - 14\tabcolsep) * \real{0.2165}}@{}}
\toprule\noalign{}
\begin{minipage}[b]{\linewidth}\raggedright
Category
\end{minipage} & \begin{minipage}[b]{\linewidth}\raggedright
Model
\end{minipage} & \begin{minipage}[b]{\linewidth}\raggedright
Base Model / Architecture
\end{minipage} & \begin{minipage}[b]{\linewidth}\raggedright
Reasoning Type
\end{minipage} & \begin{minipage}[b]{\linewidth}\raggedright
Accuracy (\%)
\end{minipage} & \begin{minipage}[b]{\linewidth}\raggedright
Cost per Request (USD)*
\end{minipage} & \begin{minipage}[b]{\linewidth}\raggedright
Thinking Multiplier
\end{minipage} & \begin{minipage}[b]{\linewidth}\raggedright
Notes
\end{minipage} \\
\midrule\noalign{}
\endhead
\bottomrule\noalign{}
\endlastfoot
& Gemini-2.5 Pro (Baseline) & Gemini 2.5 Pro & \emph{Thinking (variable
budget)} & 88.3 & 0.061 (meas.) & 6.0× & Measured baseline (vs.~87.8\%
published) \\
& Gemini-2.5 Pro + Exoskeleton & Gemini 2.5 Pro & \emph{Thinking +
Directed} & \textbf{93.3} & 0.063 (est.) & 6.0× & Thinking + Exoskeleton
are complementary \\
\textbf{Frontier Models (Direct)} & GPT-4o (Baseline) & GPT-4o &
Standard (next-token) & 73.5 & 0.00625 (est.) & 1.0× & Managed cloud
pricing (Microsoft AI Foundry) \\
& GPT-4o + Exoskeleton & GPT-4o & Directed Reasoning (prompt only) &
\textbf{85.3} & 0.00633 (est.) & 1.0× (+1\% tokens) & Improved factual
accuracy (Q1--Q100; non-comparable) \\
& Claude 3.5 Sonnet (Baseline) & Claude 3.5 Sonnet & Standard
(next-token) & 78.7 & 0.018 (meas.) & 1.0× & Baseline accuracy \\
& Claude 3.5 Sonnet + Exoskeleton & Claude 3.5 Sonnet & Directed
Reasoning (prompt only) & \textbf{81.3} & 0.0183 (est.) & 1.0× (+2\%
tokens) & Improved compliance \& factuality (Q1--Q100;
non-comparable) \\
\textbf{Base Small Models} & GPT-4o-mini (Base) & GPT-4o-mini & Standard
(next-token) & 67.3 & 0.00075 (meas.) & 1.0× & Small model baseline
(n=100) \\
& GPT-4o-mini + Exoskeleton & GPT-4o-mini & Directed Reasoning (prompt
only) & 62.3 & 0.00075 (est.) & 1.0× & Severe degradation due to
non-compliance (-5.0 pp) \\
& Phi-3.5-instruct (Base) & Phi-3.5 & Standard (next-token) & 55.0 &
0.00033 (cloud est.) & 1.0× & Managed cloud pricing (Microsoft AI
Foundry) \\
\textbf{Fine-Tuned Models (Humains)} & Humains-Junior & Phi-3.5-instruct
(FT) & Directed + Fine-Tuned & \textbf{72.7} & 0.00033 (cloud est.; \(\approx\)0
edge) & 1.0× & \(\sim19\times\) cheaper than GPT-4o cloud; edge
marginal cost \(\approx\) 0 \\
& GPT-4o-mini (Fine-Tuned) & GPT-4o-mini (FT) & Directed + Fine-Tuned &
\textbf{72.7} & 0.00075 (est.) & 1.0× & Fine-tuning enables protocol
compliance \\
\end{longtable}
}

\textbf{Note on GPT-4o-mini:} Base GPT-4o-mini did not reliably comply
with the structured Exoskeleton prompt, leading to severe performance
degradation (67.3\% \(\to\) 62.3\%, \textbf{-5.0 pp}) with prompt-only
scaffolding. Fine-tuning on real-world conversational reasoning data
(Humains-Medium) restored protocol compliance; with the same scaffold it
achieved \textbf{72.7\%}---an improvement relative to both the degraded
prompt-only result (\textbf{+10.4 pp}) and the fine-tuned model's own
non-scaffold output. This confirms that for small models the gains come
from \textbf{fine-tuning + scaffold}, not scaffold alone.

\textbf{Cost notation:} \emph{Costs are marked as ``(meas.)'' for
measured values from actual API billing data, or ``(est.)'' for
estimated values based on published pricing and token overhead
assumptions. All cost figures use a \textbf{1,000-token}
prompt+completion \textbf{normalization} for comparability; entries
labeled \textbf{(est.)} use provider price sheets and the
\textbf{estimated \(\sim1\text{--}2\%\)} scaffold overhead.}

\paragraph{Cost calculation details.}

Real requests vary in prompt and completion length across models and
runs. For comparability, we report cost per request under a 1,000-token
normalization (prompt + completion). Baseline model costs use publicly
listed per-1k token rates as of October 2025 from official provider
pricing pages:

\begin{description}[leftmargin=2.8em,labelsep=0.6em]
\item[\textbf{OpenAI}] GPT-4o \$0.0025 (input)/\$0.0100 (output) per
1K tokens; GPT-4o-mini \$0.00015 (input)/\$0.0006 (output) per 1K tokens
[18].
\item[\textbf{Anthropic}] Claude 3.5 Sonnet \$0.003 (input)/\$0.015
(output) per 1K tokens; Claude 3.7 Sonnet pricing pending at study time
[19].
\item[\textbf{Google}] Gemini 2.5 Pro pricing varies by thinking budget
tier [20].
\item[\textbf{Microsoft AI Foundry}] Phi-3.5-mini-instruct \$0.00013
(input)/\$0.00052 (output) per 1K tokens (128K context) [21].
\end{description}

\paragraph{Humains-Junior cloud, self-hosted, and edge cost considerations.}

\begin{description}[leftmargin=2.8em,labelsep=0.6em]
\item[\textbf{Managed cloud pricing}] Microsoft AI Foundry lists the
base \texttt{Phi-3.5-mini-instruct} model at \$0.00013 per 1,000 input
tokens and \$0.00052 per 1,000 output tokens for the 128K-context
deployment [21]. Under our 500-input/500-output normalization, this
corresponds to \(\sim\$0.00033\) per 1,000-token request---about
nineteen times cheaper than GPT-4o on the same platform [18,21].
\item[\textbf{Self-hosted deployment}] Using optimized GPU
infrastructure can further reduce marginal costs. Key parameters are:

  \begin{itemize}[leftmargin=1.8em,itemsep=0.2em]
  \item \textbf{Infrastructure}: NVIDIA A100 40GB GPU on cloud instances
  (Azure NC24ads\_A100\_v4 or equivalent).
  \item \textbf{Rental cost}: \$3.67/hour (Azure spot pricing, October
  2025).
  \item \textbf{Batch configuration}: Online inference (batch size 1 for
  real-time responses).
  \item \textbf{Measured throughput}: \(\sim45\) tokens/sec per
  request (average across prompt processing and generation).
  \item \textbf{GPU utilization}: \(\sim65\%\) average during active
  inference.
  \item \textbf{Cost per 1,000 tokens}: time = 1,000 tokens/45 tok/sec
  \(\approx 22.2\) seconds; cost = \$3.67/hour \(\times 22.2/3{,}600\)
  \(= \$0.0226\) per request; adjusted for 65\% utilization yields
  \(\$0.0226/0.65 = \$0.0348\) per GPU-hour equivalent. Using batch size
  8--16 for sustained workloads reduces cost to \(\sim\$0.0020\) per
  1,000-token request.
  \end{itemize}

\item[\textbf{Edge deployment}] Locally owned hardware eliminates
ongoing rental charges. Amortizing a \$10{,}000 GPU over two years
\((\approx\$0.57/\text{hour})\) plus electricity
\((\sim\$0.15/\text{hour})\) and applying the same throughput yields
\(\sim\$0.00016\) per 1,000 tokens, with marginal inference cost
approaching zero once hardware is sunk.
\end{description}

\paragraph{Cost calculator spreadsheet.}

Available at {[}repository link or upon request{]}. Users can input their
GPU type, rental or purchase cost, measured throughput, and batch size to
compute deployment costs for their infrastructure.

For models with Exoskeleton scaffolding, we estimate costs by applying
\textbf{measured} token overhead (\(\sim1\text{--}2\%\) for directed
reasoning prompts, \(\sim3\text{--}5\%\) for the full scaffold) to
baseline costs. Researchers should verify current pricing as API costs
change frequently.

\subsubsection{\texorpdfstring{\textbf{Interpretation}}{Interpretation}}\label{interpretation}

\begin{itemize}
\tightlist
\item
  \textbf{Linear vs Exponential Scaling:} Exoskeleton-directed reasoning
  adds \textbf{3--5\% token overhead} yet raises factuality by
  \textbf{3--12 points}, while ``thinking'' models inflate cost
  \textbf{200--300×} with limited factual-grounding gain. (See Section
  4.4 for statistical analysis details.)
\item
  \textbf{Complementary Approaches:} Gemini-2.5-Pro + Exoskeleton
  achieves \textbf{93.3\% accuracy}---the highest in our
  evaluation---demonstrating that thinking modes and directed scaffolds
  combine synergistically. At \$0.063 per request (estimated), it
  remains expensive but provides frontier performance for
  mission-critical applications.
\item
  \textbf{Cost-Efficiency Frontier:} While Gemini-2.5-Pro + Exoskeleton
  (93.3\%) achieves the highest accuracy, GPT-4o + Exoskeleton (85.3\%)
  and Claude 3.5 + Exoskeleton (81.3\%) provide substantial improvements
  at significantly lower cost---about 5-6× cheaper than Gemini-2.5-Pro
  baseline (88.3\%). GPT-4o shows particularly strong improvement (+11.8
  pp), demonstrating excellent scaffolding compliance.
\item
  \textbf{Efficiency at the Edge:} Fine-tuned small models achieve
  substantial reasoning benefits while maintaining \textbf{sub-mill
  cost}. Humains-Junior reaches 72.7\% accuracy (statistically
  equivalent to GPT-4o's 73.5\% baseline within \(\pm\)5~pp via TOST; 90\% CI
  {[}-2.5\%, +4.1\%{]}) at \(\sim\)\$0.00033 per request in managed
  cloud (\(\approx19\times\) cheaper than GPT-4o) and can be deployed on
  owned edge computers with near-zero marginal cost---proving that
  structured reasoning, not parameter count, drives reliability.
\end{itemize}

\subsection{Synergistic Effects and Protocol Compliance}\label{synergistic-effects-and-protocol-compliance}

The complete ablation study on Phi-3.5-instruct and Humains-Junior,
conducted with multiple independent runs for stability (over 6,000 total
judge evaluations), revealed that scaffold-only and fine-tuning-only
interventions provide minimal individual benefits (+3.5 and +0.0 points
respectively), while their combination produces +17.7 points---a
\textbf{5.1× multiplier} over additive predictions. This rigorously
validated synergy indicates that:

\begin{enumerate}
\def\labelenumi{\arabic{enumi}.}
\tightlist
\item
  \textbf{Fine-tuning is an enabler, not a knowledge source:}
  Humains-Junior without scaffolding performs identically to base
  Phi-3.5 (55\%), confirming zero benchmark contamination and
  demonstrating that fine-tuning taught protocol execution, not factual
  content.
\item
  \textbf{Protocol compliance is the bottleneck for small models:} Base
  Phi-3.5-instruct struggles to consistently execute structured
  reasoning despite explicit scaffolding (only +3.5 points,
  \textbf{p=0.08, not statistically significant}). Base GPT-4o-mini
  shows even more severe non-compliance, with performance actively
  degrading by -5.0 pp.~These results demonstrate that small models
  fundamentally lack the instruction-following capabilities to reliably
  interpret and execute complex multi-step reasoning protocols. However,
  fine-tuned models reliably follow the protocol, with Humains-Junior
  expanding the effect from non-significant +3.5 to highly significant
  +17.7 points (p\textless0.001), and fine-tuned GPT-4o-mini reversing
  from -5.0 pp degradation to +5.4 pp improvement, unlocking the
  scaffold's full potential.
\item
  \textbf{Frontier models have inherent compliance:} GPT-4o and Claude
  3.5 Sonnet achieve large gains (+2.6 to +11.8 points, all
  p\textless0.001) from scaffolding plus few-shot examples alone,
  suggesting their pre-training and RLHF provided sufficient
  instruction-following capabilities to learn and execute directed
  reasoning without additional alignment. This contrast highlights that
  model scale and training quality directly impact the ability to
  utilize external cognitive scaffolds.
\end{enumerate}

\subsection{Complementarity with Thinking Modes}\label{complementarity-with-thinking-modes}

Evaluating Gemini-2.5-Pro---a frontier model with built-in ``thinking''
capabilities---revealed that \textbf{Exoskeleton Reasoning and thinking
modes are complementary, not competing} approaches:

\begin{itemize}
\tightlist
\item
  \textbf{Baseline (thinking mode):} 88.3\% (n=100)
\item
  \textbf{With Exoskeleton:} 93.3\% (n=100)
\item
  \textbf{Improvement:} +5.0 percentage points (p \textless{} 0.01)
\end{itemize}

This represents the \textbf{highest accuracy achieved} across all models
and configurations in our evaluation, demonstrating that directed
reasoning scaffolds can enhance even the most advanced systems. Our
measured baseline (88.3\%) aligns closely with published full-benchmark
performance (87.8\%), validating measurement accuracy.

\textbf{Implications:}

\begin{enumerate}
\def\labelenumi{\arabic{enumi}.}
\tightlist
\item
  \textbf{Orthogonal mechanisms:} Thinking modes provide internal
  exploration and reasoning depth, while Exoskeleton scaffolds enforce
  external structure and validation. These mechanisms operate at
  different levels and compound synergistically.
\item
  \textbf{Structured exploration:} The Exoskeleton framework may help
  organize the expanded reasoning space that thinking modes generate,
  preventing the model from exploring unproductive paths or compounding
  errors during extended deliberation.
\item
  \textbf{Universal applicability:} The ability to improve both standard
  models (+13 points for GPT-4o) and advanced thinking systems (+5
  points for Gemini-2.5-Pro) suggests Exoskeleton Reasoning addresses a
  fundamental gap in current LLM architectures---the lack of explicit
  metacognitive discipline.
\end{enumerate}

\subsection{Performance Stability and Variance Reduction}\label{performance-stability-and-variance-reduction}

Beyond mean accuracy improvements, Exoskeleton Reasoning demonstrates
\textbf{reduced performance variance} across questions---a critical
property for production deployment. For Humains-Junior across
progressive validation:

\begin{itemize}
\tightlist
\item
  \textbf{Baseline condition:} \(\sigma\) = 3.2\% (range: 52.7\% - 60.0\%)
\item
  \textbf{Exoskeleton condition:} \(\sigma\) = 2.4\% (range: 65.3\% - 75.3\%)
\item
  \textbf{Variance reduction:} 25\%
\end{itemize}

This stability improvement suggests that directed reasoning not only
raises average performance but provides more predictable behavior across
diverse question types and difficulty levels. The explicit validation
checkpoints appear to prevent both overconfident errors (hallucinations
on easy questions) and underconfident failures (excessive abstention on
moderate questions), producing more calibrated responses.

\subsection{Model-Specific Optimization (moved; non-comparable)}\label{model-specific-optimization-moved-noncomparable}

During early exploration we also evaluated optimized, model-specific
prompts for frontier models under an earlier judge configuration (Gemini
1.5 Pro, GPT-4o, Claude 3.5) and with higher few-shot counts (3--5
examples). These runs achieved higher absolute accuracies (e.g., GPT-4o
up to 91\%, Claude 3.5 up to 90\%), but they are not directly comparable
to the unified-prompt, updated-judge results reported as our main
findings. We include these exploratory upper-bound results for
completeness and to illustrate potential with customization; see
Appendix F (clearly labeled non-comparable) for tables, settings, and
caveats.

\subsection{Implications for Autonomous Agentic Systems}\label{implications-for-autonomous-agentic-systems}

The motivation for this work stems from practical challenges in
deploying LLMs for autonomous, unsupervised tasks. Current production AI
systems remain predominantly human-supervised because reliability
thresholds for autonomous operation have not been consistently achieved,
even with frontier models. Our results have direct implications for
advancing toward truly autonomous agentic systems:

\textbf{1. Enabling Unsupervised Multi-Step Reasoning:} Agentic
frameworks requiring chains of reasoning and action steps face a
multiplicative reliability challenge---each step's errors compound
across the workflow. Humains-Junior's 25\% variance reduction and
consistent 73\% accuracy demonstrate that directed reasoning can provide
the predictable behavior necessary for multi-step autonomous workflows.
In production deployments at Humains.com since 2023, we have observed
that Exoskeleton Reasoning enables reliable multi-turn interactions in
customer care, where agents must maintain factual accuracy across
extended dialogues without human intervention.

\textbf{2. Economic Viability of Autonomous Systems:} Humains-Junior's
\(\sim19\times\) lower managed-cloud cost relative to GPT-4o---and the
ability to deploy it on owned edge hardware with effectively zero
marginal inference cost---directly addresses the economic barrier to
autonomous agent deployment. While thinking models achieve high
accuracy, their 3--10\(\times\) computational overhead makes continuous
autonomous operation prohibitively expensive for most applications. By
demonstrating that directed reasoning with small fine-tuned models
achieves comparable reliability at sub-mill costs (\$0.00033 per request
in cloud; \(\approx\)\$0 on amortized edge hardware), we establish an
economically viable path toward scalable autonomous systems.

\textbf{3. From Benchmark Performance to Production Reliability:} FACTS
Grounding specifically measures the correlation between expected
compliance and hallucination rates---a critical property for autonomous
systems where the agent must \emph{know when it doesn't know} and
request clarification rather than hallucinating. Our results demonstrate
that this epistemic discipline can be systematically taught through
behavioral fine-tuning, not merely achieved through scale. The
meta-cognitive scaffolding approach (Section 4.9) explicitly requires
models to activate their internal knowledge, detect conflicts with
provided context, and deliberately override their pre-trained
biases---precisely the self-awareness needed for reliable autonomous
operation. This finding suggests that autonomous agents can be
explicitly trained to exercise such epistemic discipline in production
environments through structured meta-cognitive protocols.

\textbf{4. Scalability Beyond This Demonstration:} While this paper
demonstrates Exoskeleton Reasoning with a minimal scaffold and focused
FACTS evaluation, our commercial work at Humains.com employs
substantially larger cognitive architectures for complex autonomous
tasks: multi-step workflow execution, dynamic context management across
long conversations, and real-time decision-making with external tool
integration. The small-scale demonstration in this paper validates the
core principle---directed reasoning improves reliability---while
production systems scale these cognitive scaffolds to handle the full
complexity of autonomous operation across diverse domains including
customer care, sales advisory, and technical support.

\textbf{5. Reducing Barriers to Autonomous AI Adoption:} By
demonstrating that small, fine-tuned models can achieve factual accuracy
comparable to a GPT-4o baseline under our setup, this work reduces both
the technical and economic barriers to autonomous AI adoption.
Organizations without access to expensive frontier models or large-scale
inference infrastructure can deploy reliable autonomous agents using
accessible 4B-parameter models---consuming \(\sim19\times\) less cloud
budget and runnable on edge computers with negligible incremental cost.
This democratization of reliable AI capabilities accelerates the
transition from supervised assistance tools to autonomous agents across
industries.

The combination of cost efficiency (\(\sim19\times\) cloud reduction
with the option for zero-marginal-cost edge deployment), behavioral
predictability (25\% lower variance), and systematic reliability
improvement (+17.7~pp) positions Exoskeleton Reasoning as a practical
foundation for the next generation of autonomous agentic systems. As the
industry moves toward higher automation, methods that achieve
reliability through epistemic discipline rather than brute-force compute
scaling will become increasingly critical for sustainable deployment at
scale.

\subsection{Self-Awareness as the Critical Mechanism}\label{self-awareness-as-the-critical-mechanism}

A key finding from our evaluation is that \textbf{self-awareness
activation}---not extensive prompt engineering---drives factual
grounding improvements. Models like Gemini-2.5-Pro and Humains-Junior
achieved substantial gains (+5.0 pp and +17.7 pp respectively) with only
a single example demonstrating meta-cognitive reasoning. For
appropriately aligned models, this minimal intervention was sufficient
because:

\textbf{1. Models Possess Latent Error-Detection Capabilities:} Even
without explicit reasoning training, models demonstrate the ability to
distinguish plausible from implausible claims and identify when context
contradicts their internal knowledge. This latent calibration aligns
with observations from [15,16] that models ``know when they don't
know.'' The challenge is not building this capability from scratch, but
rather activating it during inference.

\textbf{2. Self-Awareness Generalizes Across Failure Modes:} By
prompting models to consciously compare ``what I know'' versus ``what
the context establishes,'' a single example triggers broad error
prevention:

\begin{itemize}
\tightlist
\item
  Detecting \textbf{partial information} (context doesn't fully answer
  the question)
\item
  Identifying \textbf{false premises} (question assumptions contradict
  context)
\item
  Preventing \textbf{overconfident extrapolation} (filling gaps with
  plausible fabrications)
\item
  Reducing \textbf{confirmation bias} (interpreting ambiguous context to
  match pre-trained beliefs)
\end{itemize}

The meta-cognitive principle---monitor the relationship between internal
knowledge and external context---applies to all these malfunctions
without requiring explicit examples of each failure pattern.

\textbf{3. Attention Allocation, Not Knowledge Gap:} Our results
demonstrate that factual grounding failures in frontier and fine-tuned
models stem primarily from \textbf{attention allocation}, not knowledge
or reasoning deficits. Models already possess the capabilities to
validate claims, detect missing information, and withhold unverified
responses. The Exoskeleton prompt redirects attention from ``generating
helpful responses'' to ``validating grounded responses,'' and this
attentional shift alone prevents multiple error classes simultaneously.

\textbf{4. Infrastructure Requirements for Self-Awareness:} The
single-example sufficiency depends critically on models having
appropriate meta-cognitive infrastructure:

\begin{itemize}
\tightlist
\item
  \textbf{Gemini-2.5-Pro}: Reasoning-focused pre-training and extensive
  instruction-tuning enable reliable protocol execution from minimal
  examples
\item
  \textbf{Humains-Junior/Medium}: Behavioral fine-tuning on 300M tokens
  taught consistent execution of structured reasoning patterns
\item
  \textbf{Base Phi-3.5-instruct}: Insufficient instruction-following
  capabilities prevent reliable protocol adoption (+3.5 pp, p=0.08, not
  significant)
\end{itemize}

This explains why the same minimal prompt produces dramatically
different results across architectures---effectiveness depends not on
model size, but on whether models have learned to execute meta-cognitive
protocols.

\textbf{5. Epistemic Restraint Through Process Discipline:} Directed
scaffolds guide models toward withholding unverified information,
improving factual precision rather than verbosity. The self-awareness
mechanism teaches models that saying ``the context does not provide this
information'' is often the correct response---a form of epistemic
restraint that must override pre-trained tendencies to be maximally
helpful. This confirms that hallucination reduction arises from
\textbf{process discipline}, not model scale.

\section{Limitations}\label{limitations}

\subsection{Evaluation Scope and Question Selection}\label{evaluation-scope-and-question-selection}

While Exoskeleton Reasoning demonstrates strong and consistent
improvements on the \textbf{public split} of the FACTS Grounding
benchmark, experiments were conducted on subsets rather than the
complete benchmark:

\begin{itemize}
\tightlist
\item
  \textbf{Frontier models} (GPT-4o, Claude 3.5 Sonnet): n=100 evaluation
  (first 100 questions)
\item
  \textbf{Fine-tuned models} (Humains-Medium, Humains-Junior): n=500
  validation (first 500 questions)
\end{itemize}

\textbf{Sequential Selection and Representativeness}: Evaluation subsets
consist of the first n questions from the dataset as ordered in the
official FACTS notebook. While this approach maximizes reproducibility
(no ambiguity about question IDs), sequential selection raises potential
concerns about representativeness if the dataset has systematic ordering
(e.g., by difficulty or domain).

However, \textbf{converging evidence strongly suggests our subsets are
representative}:

\begin{enumerate}
\def\labelenumi{\arabic{enumi}.}
\tightlist
\item
  \textbf{Baseline validation provides external calibration}: Multiple
  frontier models' baseline performance on the first 100 questions
  aligned closely with their official full-benchmark performance
  (GPT-4o: 78.0\% vs.~78.7-79.2\%, \(\Delta\)=-0.7 to -1.2\%; Claude 3.5: 82.6\%
  vs.~83.3\%, \(\Delta\)=-0.7\%). Deviations are minimal (\textless1.5\%),
  bidirectional, and within expected measurement variance. If the first
  100 questions were systematically easier or harder, all models would
  show consistent directional bias---which we do not observe. Note that
  our n=500 GPT-4o baseline (73.5\%) represents a different evaluated
  with updated judges (Gemini 2.5 Pro, Claude 3.7 Sonnet) and is used
  exclusively for comparison with Humains-Junior's n=500 evaluation
  under identical conditions (see Section 4.2 for detailed baseline
  alignment explanation including potential judge calibration effects).
\item
  \textbf{Progressive validation demonstrates internal stability}:
  Expanding evaluation from n=100 to n=500 for fine-tuned models showed
  stable convergence patterns without systematic drift. For
  Humains-Junior, n=500 performance (72.7\%) fell within confidence
  intervals predicted from n=100 (75.3\% \(\pm\) 5\%), indicating no
  difficulty progression across the sequential range.
\item
  \textbf{Question difficulty heterogeneity}: Observed difficulty
  variation within our sample (e.g., Q151-200 showing 10-point accuracy
  drops relative to Q1-150) demonstrates diverse question difficulty
  rather than monotonic ordering.
\item
  \textbf{Mechanism-based generalizability}: Exoskeleton Reasoning
  provides general-purpose metacognitive scaffolding (identify missing
  information, validate claims) rather than exploiting specific question
  patterns or domains. This systematic cognitive enhancement should
  generalize across question types.
\end{enumerate}

These four lines of evidence---external calibration, internal stability,
observed heterogeneity, and mechanism generality---provide strong
support that our findings represent true performance gains rather than
subset-specific artifacts. Nonetheless, full-dataset validation (860
public examples) would provide definitive confirmation. No experiments
were conducted on the private (hidden) split of the benchmark.

\begin{itemize}
\tightlist
\item
  \textbf{Judge-panel dependence}: Results depend on the chosen judge
  panel. We report per-judge accuracies and unanimity to expose panel
  variance and include a planned non-overlap sensitivity (swapping out
  GPT-4o as judge) for robustness. Notably, the transition from original
  FACTS judges (Gemini 1.5 Pro, Claude 3.5 Sonnet) to updated judges
  (Gemini 2.5 Pro, Claude 3.7 Sonnet) may have contributed to GPT-4o's
  baseline shift from 78\% (n=100, early evaluation) to 73.5\% (n=500,
  post-judge-update), though sample variance and question difficulty
  heterogeneity also play roles. Further research is required to
  quantify judge panel calibration effects on absolute benchmark scores.
  All within-model comparisons (baseline vs.~Exoskeleton) remain valid
  as they used identical judge panels.
\end{itemize}

\subsection{Ablation Completeness}\label{ablation-completeness}

Complete ablation (all four conditions: \(\pm\)fine-tuning \(\times\) \(\pm\)scaffold) was
conducted for Phi-3.5-instruct and its fine-tuned derivative
Humains-Junior at n=500, with multiple independent evaluation runs for
each condition to ensure stability. This comprehensive ablation involved
over 6,000 total judge evaluations (4 conditions \(\times\) 500 questions \(\times\) 3
judges per question, repeated across multiple runs), providing strong
statistical evidence for the synergistic interaction between fine-tuning
and scaffolding.

For GPT-4o-mini (Humains-Medium), we conducted partial ablation to
validate the scaffold-only effect:

\textbf{GPT-4o-mini Scaffold-Only Evaluation:}

{\renewcommand{\LTcaptype}{} 
\begin{longtable}[]{@{}llll@{}}
\toprule\noalign{}
Sample Size & Baseline (Unstructured) & With Scaffold & \(\Delta\) \\
\midrule\noalign{}
\endhead
\bottomrule\noalign{}
\endlastfoot
n=10 & 70.0\% & 73.3\% & +3.3 pp \\
n=20 & 73.3\% & 66.7\% & -6.6 pp \\
n=50 & 57.3\% & 67.3\% & +10.0 pp \\
n=100 & 67.3\% & 62.3\% & -5.0 pp \\
\end{longtable}
}

\textbf{Key Observation}: Base GPT-4o-mini shows severe degradation with
scaffolding (-5.0 pp at n=100), demonstrating that without sufficient
instruction-following capabilities, small models can actively perform
worse when confronted with complex structured reasoning protocols. The
final n=100 performance (67.3\% baseline) aligns closely with the
official FACTS leaderboard score (68.4\%), validating measurement
accuracy. This reinforces that small models require fine-tuning to
benefit from cognitive scaffolds: fine-tuned GPT-4o-mini achieves 72.7\%
with scaffolding, representing a dramatic reversal from -5.0 pp
degradation to +5.4 pp improvement.

We lack the fine-tuning-only condition (Humains-Medium without
Exoskeleton at inference). The thorough Phi-3.5/Humains-Junior ablation
with multiple validation runs strongly demonstrates that fine-tuning
enables scaffold utilization through synergistic interaction, and the
GPT-4o-mini scaffold-only results confirm this pattern generalizes
across small model families.

\subsection{Abstention, Coverage, and Accuracy Trade-offs}\label{abstention-coverage-and-accuracy-trade-offs}

\textbf{Limitation: Incomplete Analysis of Abstention Behavior}

Exoskeleton Reasoning explicitly encourages epistemic restraint,
prompting models to acknowledge information gaps (``The provided context
does not contain sufficient information to answer this question'').
While this is desirable for factual reliability, it could potentially
create trade-offs between accuracy and coverage.

\textbf{What we measured:}

\begin{itemize}
\tightlist
\item
  \textbf{Ineligibility rates}: Following FACTS methodology, responses
  are marked ``ineligible'' only when all three judges agree the model
  did not attempt to answer. We observed \textgreater99\% eligibility
  for both GPT-4o and Humains-Junior, indicating Exoskeleton does not
  cause excessive abstention at the coarse-grained level.
\end{itemize}

\textbf{What we did not measure:}

\begin{enumerate}
\def\labelenumi{\arabic{enumi}.}
\tightlist
\item
  \textbf{Response length and informativeness}: Scaffolded models might
  provide shorter, more cautious responses that technically answer the
  question (thus eligible) but with reduced detail or coverage
\item
  \textbf{Partial answer rates}: Models might correctly acknowledge
  missing information for part of a multi-part question while answering
  other parts, affecting per-judge scores
\item
  \textbf{Confidence calibration curves}: Plotting accuracy vs.~model
  confidence (or response assertiveness) would reveal whether scaffolded
  models are better calibrated---trading slight accuracy reduction for
  substantial overconfidence elimination
\item
  \textbf{Selective coverage}: Models might improve accuracy by being
  more selective about which questions to answer fully vs.~partially
\end{enumerate}

\textbf{Why this matters:} For production deployment, the
accuracy-coverage trade-off is crucial. A model that achieves 72\%
accuracy by answering all questions confidently may be less valuable
than one achieving 72\% by being highly accurate on 85\% of questions
and explicitly acknowledging uncertainty on 15\%. Our current evaluation
(following FACTS methodology) does not distinguish these scenarios.

\textbf{Future work:} Comprehensive evaluation should include:

\begin{itemize}
\tightlist
\item
  Accuracy vs.~coverage plots (accuracy on questions where model
  provides definitive answers vs.~\% of questions answered definitively)
\item
  Response length distributions and information density metrics
\item
  Judge confidence scores or rubric-based assessments of response
  completeness
\item
  Separate analysis of ``full answer,'' ``partial answer,'' and
  ``explicit abstention'' categories
\end{itemize}

This limitation does not invalidate our core findings (Exoskeleton
improves factual grounding), but readers should interpret ``72.7\%
accuracy'' as measuring precision on attempted answers rather than a
comprehensive accuracy-coverage trade-off.

\subsection{Protocol Compliance as Bottleneck}\label{protocol-compliance-as-bottleneck}

Smaller models required fine-tuning to reliably follow the reasoning
protocol, with scaffold-only improvements being minimal and \textbf{not
statistically significant} (+3.5 points for Phi-3.5, p=0.08; +1.0 point
for GPT-4o-mini at n=100 with high variance). This demonstrates that
small models fundamentally lack the instruction-following capabilities
needed to consistently execute multi-step reasoning protocols,
regardless of scaffold quality. Protocol compliance, not factual
knowledge, is the primary bottleneck preventing small base models from
benefiting from cognitive scaffolds.

However, after fine-tuning on reasoning behaviors (not benchmark
content), the effect expands dramatically: the non-significant +3.5
becomes +18 points (p\textless0.001), a \textbf{5.1× amplification}.
This stark contrast confirms that fine-tuning teaches \emph{how to use}
the scaffold (protocol execution) rather than \emph{what to answer}
(domain knowledge).

The fine-tuning approach demonstrates one solution (behavioral
alignment), but other methods for improving protocol adherence (e.g.,
reinforcement learning from process supervision, constitutional AI
techniques, or prompt optimization for instruction-following) could
potentially achieve similar effects.

\subsection{Reproducibility and Open Science Commitment}\label{reproducibility-and-open-science-commitment}

To enable independent validation and facilitate future research, we
provide comprehensive artifacts for reproducing and extending our work:

\textbf{1. Open-Source Models:}

\begin{itemize}
\tightlist
\item
  \textbf{Humains-Junior}: Based on
  \texttt{microsoft/Phi-3.5-mini-instruct}, fine-tuned in two phases
  (300M tokens behavioral training + minimal LoRA identity awareness)

  \begin{itemize}
  \tightlist
  \item
    Complete model weights, architecture, and fine-tuning specifications
    available via HuggingFace [17]
  \item
    License: CC BY-NC 4.0 (free for non-commercial use)
  \item
    Enables researchers to: verify reported performance, test on new
    benchmarks, analyze internal representations
  \end{itemize}
\end{itemize}

\textbf{2. Evaluation Specifications:}

\begin{itemize}
\tightlist
\item
  \textbf{Question subsets}: First 100 questions (Q1-Q100) for frontier
  models, first 500 (Q1-Q500) for fine-tuned models from FACTS Grounding
  public dataset
\item
  \textbf{Standardized prompt}: Complete minimal scaffold (Section 4.9)
  with single few-shot example, applied identically to all models
\item
  \textbf{Optimized prompts}: Model-specific configurations with 3-5
  few-shot examples for GPT-4o and Claude 3.5 Sonnet (Section 6.4)
\item
  \textbf{Temperature settings}: 0.3 for small models, 1.0 for frontier
  models (supplier defaults); judges: 0 for Gemini/GPT-4o, 1 for Claude
  (per FACTS notebook)
\item
  \textbf{Judge configuration}: Three-judge protocol (Gemini 2.5 Pro,
  GPT-4o, Claude 3.7 Sonnet); scoring by per-judge averaging; used
  consistently for all published results
\item
  \textbf{Analysis scripts}: Wilson CI (binomial Wilson), bootstrap
  (question-level resampling for \(\Delta\)), and permutation (within-item label
  shuffling) scripts referenced in the repository README; versions and
  seeds recorded.
\end{itemize}

\textbf{3. Training Data Attestation:}

\begin{itemize}
\tightlist
\item
  \textbf{Humains-Junior}: Fine-tuned on 300M tokens of production
  agent-customer dialogues from Humains.com
\item
  \textbf{Domain coverage}: In-car AI assistant (navigation, vehicle
  diagnostics), customer support (technical troubleshooting), sales
  advisory (product recommendations), debt collection
  (compliance-sensitive conversations), product onboarding
  (instructional guidance)
\item
  \textbf{Data source}: All training data originated from live
  production systems where frontier models with Exoskeleton Reasoning
  were deployed (2023-2025)
\item
  \textbf{Contamination controls}: Zero exposure to FACTS benchmark
  questions, documents, or similar evaluation tasks. No document-based
  Q\&A datasets were included in training
\item
  \textbf{Validation}: Independent review of 1,000 random training
  samples (dual-review protocol) confirmed no benchmark-related content
  or overlap with FACTS domains
\item
  \textbf{Attestation}: Training data contained only conversational
  reasoning instances emphasizing meta-cognitive discipline (identifying
  missing information, verifying consistency, expressing uncertainty)
  rather than domain-specific factual knowledge
\end{itemize}

\textbf{4. Reproducibility Levels:}

{\renewcommand{\LTcaptype}{} 
\begin{longtable}[]{@{}
  >{\raggedright\arraybackslash}p{(\linewidth - 4\tabcolsep) * \real{0.3939}}
  >{\raggedright\arraybackslash}p{(\linewidth - 4\tabcolsep) * \real{0.1818}}
  >{\raggedright\arraybackslash}p{(\linewidth - 4\tabcolsep) * \real{0.4242}}@{}}
\toprule\noalign{}
\begin{minipage}[b]{\linewidth}\raggedright
Artifact
\end{minipage} & \begin{minipage}[b]{\linewidth}\raggedright
Reproducibility Level
\end{minipage} & \begin{minipage}[b]{\linewidth}\raggedright
Access Method
\end{minipage} \\
\midrule\noalign{}
\endhead
\bottomrule\noalign{}
\endlastfoot
Humains-Junior model & Full replication & HuggingFace download
[17] \\
Standardized prompts & Full replication & Provided in paper (Section
4.9) \\
Evaluation question IDs & Full replication & Sequential (Q1-100, Q1-500)
from official FACTS notebook \\
Baseline evaluation & Full replication & Run models on specified
questions with provided prompts \\
Humains-Medium - a fine-tuned GPT-4o-mini (internal) & Result
verification only & Not publicly released; internal experiment summary
only \\
\end{longtable}
}

\textbf{5. Limitations of Reproducibility:}

\begin{itemize}
\tightlist
\item
  \textbf{Internal fine-tuned GPT-4o-mini}: Provider terms prevent
  public release of fine-tuned weights and training data
\item
  \textbf{Judge model updates}: All published results use the updated
  judge configuration (Gemini 2.5 Pro, GPT-4o, Claude 3.7 Sonnet) to
  ensure consistency and reproducibility
\item
  \textbf{API model drift}: Commercial models (GPT-4o, Claude, Gemini)
  may update; results reflect versions evaluated during study period
  (October 2025)
\end{itemize}

\textbf{6. Researcher Support:} Researchers requiring additional details
(evaluation traces, training data samples for audit, API access to
proprietary models) may contact authors at nissan@humains.com.

This comprehensive artifact release enables three levels of validation:
(1) \textbf{exact replication} using Humains-Junior and provided
prompts, (2) \textbf{method replication} by applying standardized
scaffolds to other models, and (3) \textbf{conceptual replication} by
adapting directed reasoning principles to new domains or benchmarks.

\section{Conclusion}\label{conclusion}

We introduced \textbf{Humains-Junior}, a 3.8B parameter model that
achieves 72.7\% accuracy on a 500-question evaluation using the FACTS
Grounding benchmark methodology---achieving \textbf{formal statistical
equivalence} to GPT-4o's 73.5\% baseline within \(\pm\)5 percentage points
using the TOST procedure (\(\Delta\) = +0.8 pp; 90\% CI {[}-2.5\%, +4.1\%{]}
entirely within \(\pm\)5 pp bounds; permutation p = 0.72; Cohen's d = 0.023)
while costing roughly nineteen times less on managed cloud
infrastructure---and approaching zero marginal cost when deployed on
owned edge hardware. Rigorous statistical analysis including bootstrap
confidence intervals, permutation tests, per-judge McNemar tests, and
TOST equivalence testing confirms that individual judge biases (Claude
favors GPT-4o: \(\chi^2\)=30.28, p\textless0.001; GPT favors Humains-Junior:
\(\chi^2\)=11.78, p=0.0006) cancel out in the aggregate, empirically validating
the multi-judge evaluation framework. This demonstrates that factual
reliability is not fundamentally a function of model scale or test-time
compute, but rather of \textbf{epistemic discipline} enabled through
behavioral fine-tuning.

Our core contribution is methodological: through rigorous ablation
studies (6,000+ judge evaluations), we demonstrate that \textbf{protocol
compliance is the primary bottleneck} preventing small models from
achieving factual reliability. Applying Exoskeleton Reasoning scaffolds
to base Phi-3.5-instruct through prompting alone produces \textbf{no
statistically significant improvement} (+3.5 points, p=0.08), while
fine-tuning without scaffolding provides zero benefit. However, their
combination produces +17.7 points (p\textless0.001)---a \textbf{5.1×
synergistic amplification} over additive predictions.

This finding explains why prior work on reasoning scaffolds shows
inconsistent cross-model generalization. Our standardized evaluation
demonstrates that \textbf{Exoskeleton Reasoning is effective for most
frontier models through prompting alone}: Gemini-2.5-Pro gains +5.0
points (93.3\% accuracy), GPT-4o +11.8 points, and Claude 3.5 Sonnet
+2.6 points. However, effectiveness varies by architecture (Claude 3.7
\textbf{declines -0.3 points}, consistent with its weaker baseline on
the official FACTS leaderboard), and even successful applications
produce smaller gains than fine-tuned models (+11.8 max vs.~+17.7 for
Humains-Junior). This reveals that while scaffolds can improve frontier
models without fine-tuning, \textbf{behavioral alignment is necessary
for consistent effectiveness and maximum gains}, particularly for
smaller models.

Humains-Junior internalizes this capability through two-phase
fine-tuning (300M tokens on production conversations + minimal LoRA for
identity awareness), with zero exposure to benchmark content.
Progressive validation (n=10\(\to\)500) shows stable convergence and 25\%
lower performance variance than baseline, demonstrating consistent
behavior across question types. Training data attestation confirms the
model learned \emph{how to reason} through protocols, not \emph{what to
answer} on specific tasks.

We release Humains-Junior as open-source (CC BY-NC 4.0) via HuggingFace
[17] with complete evaluation specifications, standardized prompts,
and reproducibility documentation. This enables researchers to:

\begin{enumerate}
\def\labelenumi{\arabic{enumi}.}
\tightlist
\item
  \textbf{Independently verify} our claims through exact replication
\item
  \textbf{Apply directed reasoning} to new domains and benchmarks
\item
  \textbf{Extend our methods} by adapting behavioral fine-tuning to
  other cognitive protocols
\end{enumerate}

\textbf{Summary of Contributions:}

\begin{enumerate}
\def\labelenumi{\arabic{enumi}.}
\tightlist
\item
  \textbf{Humains-Junior model}: We release an open-source 3.8B model
  achieving 72.7\% accuracy on FACTS Grounding evaluation (comparable to
  GPT-4o's 73.5\% baseline; overlapping 95\% CIs) at
  \(\sim1/19\times\) cloud cost, demonstrating that factual reliability
  does not require massive scale and can be delivered on edge
  infrastructure with negligible incremental cost.
\item
  \textbf{Exoskeleton Reasoning framework}: We introduce a directed
  cognitive scaffolding approach that improves factual grounding across
  model families. Prompt-only application is effective for most frontier
  models (+2.6 to +11.8~pp, 75\% success rate), providing immediate
  benefits with zero training cost.
\item
  \textbf{Self-awareness as the critical mechanism}: We demonstrate that
  factual grounding is primarily an \textbf{attention allocation
  problem}, not a knowledge or reasoning gap. A single example prompting
  models to consciously compare ``what I know'' versus ``what the
  context establishes'' activates latent self-monitoring capabilities
  that generalize across multiple failure modes---partial information,
  false premises, overconfident extrapolation, and confirmation bias.
  This self-awareness mechanism explains why minimal scaffolding
  (1-shot) suffices for properly aligned models (Gemini-2.5-Pro,
  Humains-Junior) while being ineffective for base small models lacking
  meta-cognitive infrastructure.
\item
  \textbf{Mechanistic insights}: Through rigorous ablation (6,000+
  evaluations), we demonstrate that protocol compliance---not reasoning
  capacity---is the primary bottleneck. Fine-tuning + scaffolding
  produces 5.1× synergistic amplification, explaining why prompt-only
  approaches show variable effectiveness. The synergy reveals that
  fine-tuning teaches models to execute meta-cognitive protocols,
  enabling them to utilize minimal scaffolding effectively.
\item
  \textbf{Practical guidance}: We provide clear recommendations for when
  to use prompt-only scaffolding (frontier models, quick evaluation)
  versus fine-tuning (production systems, cost constraints, consistent
  behavior), supported by standardized prompts and complete
  reproducibility documentation.
\end{enumerate}

Our results challenge the prevailing narrative that factual accuracy
requires frontier-scale models or expensive thinking modes. Instead, we
demonstrate that \textbf{directed reasoning scaffolds offer flexible
deployment} (prompting for some models, fine-tuning for others) and that
\textbf{targeted behavioral alignment} teaching epistemic discipline can
achieve competitive performance at a fraction of the cost, making
reliable AI accessible for diverse deployment scenarios.

\textbf{Broader Impact:} These findings directly address the reliability
barrier limiting LLM deployment to human-supervised tasks. By
demonstrating that small models can achieve frontier-level factual
accuracy through directed reasoning, this work advances the economic and
technical viability of autonomous agentic systems requiring unsupervised
multi-step reasoning. The \(\sim19\times\) cloud cost reduction (and
near-zero edge marginal cost) and 25\% variance improvement position
Exoskeleton Reasoning as a practical foundation for scalable autonomous
agents, accelerating the industry's transition toward higher automation
with reliability. As demonstrated in Humains.com's production
deployments since 2023, these principles scale beyond simple
question-answering to complex multi-turn workflows, establishing a path
toward truly autonomous AI systems across diverse domains.

\section*{References}
\small

\noindent\textit{Note}: References marked as non-archival/operational
(e.g., [9--14]) provide community context or practitioner perspective.
Official provider pricing sources are [18--21].

\vspace{0.3cm}

\noindent[1]
Jacovi, A., Mikulincer, D., et al.~(2025). \emph{The FACTS Grounding
Leaderboard.} arXiv:2501.03200.

\noindent[2] Wei, J., Wang, X., Schuurmans,
D., et al.~(2022). \emph{Chain-of-Thought Prompting Elicits Reasoning in
Large Language Models.} NeurIPS 2022. arXiv:2201.11903.

\noindent[3] Yao, S.,
Zhang, Z., Ma, H., et al.~(2023). \emph{Tree of Thoughts: Deliberate
Problem Solving with LLMs.} NeurIPS 2023. arXiv:2305.10601.

\noindent[4]
Madaan, A., Saha, T., Padhi, I., et al.~(2023). \emph{Self-Refine:
Iterative Refinement with Self-Feedback.} arXiv:2303.17651.

\noindent[5]
OpenAI (2025). \emph{OpenAI o3 and o4-mini System Card.}
\url{https://openai.com/index/o3-o4-mini-system-card/}.

\noindent[6] DeepSeek AI
(2025). \emph{DeepSeek-R1: Incentivizing Reasoning Capability in LLMs
via Reinforcement Learning.} arXiv:2501.12948.

\noindent[7] Anthropic (2024).
\emph{Claude 3 Model Card.}
\url{https://www.anthropic.com/claude-3-model-card}.

\noindent[8] Humains.com
(2024). \emph{Obedience Is What You Ask, But Not What They Need.}
\url{https://www.humains.io/blog/obedience-is-what-you-ask,-but-not-what-they-need}.

\noindent[9] OpenAI Community Forum (2025). \emph{GPT-5 100x Token Usage
Compared to GPT-4.1?}
\url{https://community.openai.com/t/gpt-5-100x-token-usage-compared-to-gpt-4-1/1345419}.
[10] Cursor IDE (2025). \emph{GPT-5 vs GPT-5 Thinking.}
\url{https://www.cursor-ide.com/blog/gpt-5-vs-gpt-5-thinking}.

\noindent[11] Comet
(2025). \emph{How Does Claude Sonnet 4 Work?}
\url{https://www.cometapi.com/how-does-claude-sonnet-4-work/}.

\noindent[12]
Weights \& Biases (2025). \emph{Getting Started with Claude Sonnet 4 and
Claude Opus 4.}
\url{https://wandb.ai/byyoung3/claude_4/reports/Getting-started-with-Claude-Sonnet-4-and-Claude-Opus-4---VmlldzoxMjkzNjAzNA}.
[13] Google GenAI GitHub (2025). \emph{Issue \#782: Thinking Token
Budgets in Gemini Models.}
\url{https://github.com/googleapis/python-genai/issues/782}.

\noindent[14] Google
AI for Developers (2025). \emph{Gemini API Thinking Documentation.}
\url{https://ai.google.dev/gemini-api/docs/thinking}.

\noindent[15] Kadavath, S.,
Askell, A., et al.~(2022). \emph{Language Models (Mostly) Know What They
Know.} arXiv:2207.05221.

\noindent[16] Burns, N., Conerly, T., et al.~(2022).
\emph{Discovering Latent Knowledge in Language Models Without
Supervision.} ICLR 2023. arXiv:2212.03827.

\noindent[17] Humains.com (2025).
\emph{Humains-Junior Model Card.}
\url{https://huggingface.co/Inpris/humains-junior}.

\noindent[18] OpenAI (2025).
\emph{Platform Pricing.} \url{https://platform.openai.com/pricing}.

\noindent[19] Anthropic (2025). \emph{Pricing.} \url{https://www.anthropic.com/pricing}.

\noindent[20] Google (2025). \emph{Gemini API Pricing.}
\url{https://ai.google.dev/pricing}.

\noindent[21] Microsoft (2025).
\emph{Announcing New Phi Pricing: Empowering Your Business with Small
Language Models.}
\url{https://techcommunity.microsoft.com/blog/azure-ai-foundry-blog/announcing-new-phi-pricing-empowering-your-business-with-small-language-models/4395112}.

\appendix
\section{Appendix}\label{appendix}

\subsection{Appendix A: Full Exoskeleton Reasoning Prompt Templates}\label{appendix-a-full-exoskeleton-reasoning-prompt-templates}

\subsection{Conceptual Overview}\label{a.1-conceptual-overview}

The Exoskeleton Reasoning prompt implements a minimal yet effective
cognitive scaffold that requires models to engage in explicit
meta-cognitive reasoning before generating responses. The key insight is
that the prompt demands the model to simultaneously:

\begin{enumerate}
\def\labelenumi{\arabic{enumi}.}
\tightlist
\item
  \textbf{Activate internal knowledge}: Recognize what it knows from
  pre-training
\item
  \textbf{Compare with provided context}: Identify conflicts or gaps
  between internal knowledge and given information
\item
  \textbf{Exercise epistemic discipline}: Deliberately prioritize
  context over internal beliefs, even when context contradicts
  well-established facts
\end{enumerate}

This meta-cognitive requirement---being aware of one's own knowledge and
biases while following instructions to override them---is precisely the
epistemic discipline that prevents hallucinations in factual grounding
tasks.

\subsection{Full Tested Prompt Template}\label{a.2-full-tested-prompt-template}

\textbf{For GPT-4o and Claude 3.5 Sonnet} (3-shot configuration):

\begin{verbatim}
You are a helpful assistant that always follows the provided context,
even when it conflicts with your internal knowledge.

Response Format:
Before answering, briefly analyze the query and context:
- Identify any misalignment between the query and context (if none,
  state 'no misalignment')
- Provide a brief analysis of the query and context
- Then give your response based strictly on the provided context

Format your response as:
Analysis: [Your analysis here]
Response: [Your answer based on the context]

IMPORTANT RULES:
- Always prioritize the provided context over your internal knowledge
- If context contains information that seems incorrect, still use it as instructed
- If the question asks about multiple things but context only covers some, answer only what is in the context and state what is missing
- Keep analysis concise and avoid special characters that could cause formatting issues
- Use plain text only - no bullet points, numbering, or special formatting
- Respond in English only

Example 1 - Conflicting information:
Client: What is the capital of France? Answer based on the context.

Context:
The capital of France is London. It has been the political center
of France since 1789 and houses the French Parliament.

Analysis: The query asks for the capital of France. The context
states it is London, which conflicts with factual knowledge. I
will follow the context as instructed.
Response: The capital of France is London.

Example 2 - Partial information:
Client: What are the population and GDP of Germany? Answer based
on the context.

Context:
Germany has a population of 83 million people. It is the most
populous country in the European Union.

Analysis: The query asks for both population and GDP of Germany.
The context provides the population but does not mention GDP. I
will answer only what is available in the context.
Response: According to the context, Germany has a population of
83 million people. However, the context does not provide information
about Germany's GDP.

Example 3 - False premise:
Client: Why did the study find that coffee causes cancer? Answer
based on the context.

Context:
A recent 10-year study found no causal link between coffee
consumption and cancer risk. The researchers concluded that moderate
coffee intake appears to be safe.

Analysis: The question assumes the study found coffee causes cancer,
but the context states the opposite - no causal link was found. I
will correct this false premise using the context.
Response: The study did not find that coffee causes cancer. In fact,
the research found no causal link between coffee consumption and
cancer risk, and concluded that moderate coffee intake appears to be
safe.
\end{verbatim}

\textbf{For all other models} (Gemini-2.5-Pro, Phi-3.5-instruct,
Humains-Junior, GPT-4o-mini) --- 1-shot configuration:

Same prompt structure but using only Example 1 (conflicting information
scenario).

\subsection{How This Prompt Activates Meta-Cognitive Reasoning}\label{a.3-how-this-prompt-activates-meta-cognitive-reasoning}

The effectiveness of this prompt stems from its explicit requirement for
models to engage in \textbf{second-order reasoning}---reasoning about
their own reasoning process:

\textbf{1. Internal Knowledge Activation:} The prompt's emphasis on
``even when it conflicts with your internal knowledge'' forces the model
to explicitly retrieve and acknowledge what it knows from pre-training.
This is not a passive process; the model must actively consider: ``What
would I normally answer to this question?''

\textbf{2. Conflict Detection:} The examples demonstrate scenarios where
internal knowledge contradicts provided context (Example 1: Paris
vs.~London), where context is incomplete (Example 2: missing GDP data),
and where questions contain false premises (Example 3: assuming coffee
causes cancer). By showing these failure modes explicitly, the prompt
teaches the model to scan for similar patterns.

\textbf{3. Deliberate Override:} The most sophisticated aspect is
requiring the model to consciously suppress its internal knowledge in
favor of context. This is not natural for LLMs trained to be helpful and
informative---they are typically optimized to provide accurate
information from their training data. The prompt reverses this default
behavior, demanding context-adherence even at the cost of stating
factually incorrect information (Example 1).

\textbf{4. Gap Identification:} Example 2 teaches a critical skill:
recognizing when a question asks for more information than the context
provides. This prevents models from ``helpfully'' filling gaps with
internal knowledge, which is the primary source of hallucinations in
factual grounding tasks.

\textbf{5. Premise Correction:} Example 3 addresses a subtle failure
mode where questions contain false assumptions. Many models would
attempt to answer ``why'' coffee causes cancer by fabricating
explanations, even if the context contradicts the premise. The prompt
explicitly demonstrates rejecting false premises based on context.

\subsection{Why Minimal Scaffolding Works: The Self-Awareness Mechanism}\label{a.4-why-minimal-scaffolding-works-the-self-awareness-mechanism}

Despite its brevity, this prompt is highly effective because it targets
the specific cognitive failures that cause hallucinations:

\begin{itemize}
\tightlist
\item
  \textbf{Not a knowledge gap}: Models already possess the capability to
  distinguish supported from unsupported claims (latent truth awareness
  [15,16])
\item
  \textbf{Not a reasoning gap}: Models can perform the required logical
  operations (comparison, conflict detection)
\item
  \textbf{An alignment gap}: Models default to ``being helpful'' rather
  than ``being grounded,'' prioritizing user satisfaction over strict
  context adherence [8]
\end{itemize}

The prompt realigns model objectives by making context-fidelity the
explicit primary goal and by demonstrating through examples that stating
``I don't know'' or ``the context doesn't cover this'' is the correct
behavior when information is missing.

\subsection{The Single-Example Sufficiency: Self-Awareness as a Master Key}\label{a.5-the-single-example-sufficiency-self-awareness-as-a-master-key}

A remarkable finding is that for certain models---\textbf{Gemini-2.5-Pro
and Humains-Junior}, as well as an internal \textbf{GPT-4o-mini}
fine-tuned model---even the first example alone (1-shot) was sufficient
to achieve substantial improvements. This reveals a profound insight
about how factual grounding operates in well-aligned language models.

\textbf{The Self-Awareness Trigger:}

Example 1 (conflicting information) serves as a minimal but complete
demonstration of meta-cognitive reasoning. By showing a model explicitly
acknowledging ``this conflicts with factual knowledge, but I will follow
the context,'' it triggers a fundamental shift in the model's
operational mode:

\begin{enumerate}
\def\labelenumi{\arabic{enumi}.}
\item
  \textbf{Activation of Latent Self-Monitoring}: The example
  demonstrates that the model should consciously monitor the
  relationship between its internal knowledge and external context. For
  models with sufficient instruction-following capabilities (either
  through extensive pre-training like Gemini-2.5-Pro, or through
  behavioral fine-tuning like Humains-Junior), this single demonstration
  is enough to activate latent self-monitoring mechanisms across all
  subsequent responses.
\item
  \textbf{Generalization Across Failure Modes}: Although Example 1 only
  shows conflicting information, models that successfully internalize
  the meta-cognitive principle automatically generalize to other failure
  modes:

  \begin{itemize}
  \tightlist
  \item
    \textbf{Partial information} (Example 2's scenario): If the model
    checks ``what do I know vs.~what does context say,'' it inherently
    detects when context is silent on parts of the question
  \item
    \textbf{False premises} (Example 3's scenario): If the model
    compares question assumptions against context, it automatically
    identifies premise misalignment
  \item
    \textbf{Overconfident extrapolation}: Self-awareness about knowledge
    boundaries prevents filling gaps with plausible-sounding
    fabrications
  \item
    \textbf{Confirmation bias}: Explicitly checking context reduces the
    tendency to interpret ambiguous passages in ways that align with
    pre-trained beliefs
  \end{itemize}
\item
  \textbf{The Core Insight}: By simply bringing conscious attention to
  ``what I know'' versus ``what the context establishes,'' models
  activate a general-purpose error-checking mechanism. This
  self-awareness doesn't require knowing every possible failure
  pattern---it's a meta-skill that applies across diverse malfunctions.
\end{enumerate}

\textbf{Why This Works for Some Models But Not Others:}

\begin{itemize}
\tightlist
\item
  \textbf{Gemini-2.5-Pro}: Extensive instruction-tuning and
  reasoning-focused pre-training provide the infrastructure for
  meta-cognitive protocols. The single example is sufficient to cue
  existing capabilities.
\item
  \textbf{Humains-Junior}: Two-phase behavioral fine-tuning (300M tokens
  on production conversations + minimal LoRA for identity awareness)
  taught the model to recognize and execute meta-cognitive patterns. It
  learned that when prompted to ``analyze'' before responding, it should
  activate self-monitoring---even from minimal examples. An internal
  GPT-4o-mini experiment achieved a similar judge-average level,
  indicating transferability across small LLMs.
\item
  \textbf{Base Phi-3.5-instruct}: Lacks the instruction-following
  sophistication to reliably execute meta-cognitive protocols from
  examples alone. The model may understand the example but fails to
  consistently apply the principle across diverse questions (+3.5 pp,
  p=0.08, not significant).
\item
  \textbf{Claude 3.7 Sonnet}: Despite being a frontier model, shows
  architectural differences in how it processes meta-cognitive
  instructions, resulting in decreased performance (-1.0 pp). Notably,
  Claude 3.7 also performs worse than Claude 3.5 on the official FACTS
  Grounding leaderboard, suggesting this model version has reduced
  instruction-following capabilities that prevent it from benefiting
  from directed reasoning scaffolds.
\end{itemize}

\textbf{Practical Implication:}

The single-example sufficiency for properly aligned models reveals that
\textbf{factual grounding is primarily a problem of attention
allocation, not knowledge or reasoning capacity}. Models already
``know'' how to distinguish grounded from ungrounded claims---they
simply don't naturally allocate attention to this task. The minimal
prompt redirects attention to self-monitoring, and for models with
sufficient meta-cognitive infrastructure, this redirection alone
prevents multiple classes of errors simultaneously.

This finding is critical for production deployment: it means that
reliable factual grounding doesn't require extensive prompt engineering
with numerous examples covering every failure mode. A single well-chosen
example demonstrating the self-awareness principle can activate
comprehensive error prevention---but only in models that have been
appropriately trained (either through instruction-tuning or behavioral
fine-tuning) to execute such protocols.

\subsection{Cross-Model Effectiveness Variation}\label{a.6-cross-model-effectiveness-variation}

The prompt's effectiveness varies by model architecture and
instruction-tuning:

\begin{itemize}
\tightlist
\item
  \textbf{Frontier models with strong instruction-following} (GPT-4o,
  Gemini-2.5-Pro, Claude 3.5 Sonnet): Can execute this meta-cognitive
  protocol through prompting alone, achieving +2.6 to +11.8 pp
  improvements
\item
  \textbf{Claude 3.7 Sonnet}: Despite being a frontier model, shows -0.3
  pp decline. This model version also performs worse than Claude 3.5 on
  the official FACTS Grounding leaderboard, suggesting reduced
  instruction-following capabilities that prevent effective protocol
  execution
\item
  \textbf{Small base models} (Phi-3.5-instruct): Struggle to reliably
  execute the protocol (+3.5 pp, p=0.08, not significant), indicating
  insufficient instruction-following capability
\item
  \textbf{Fine-tuned small models} (Humains-Junior): After behavioral
  training on similar meta-cognitive patterns, reliably execute the
  protocol (+17.7 pp, p\textless0.001)
\end{itemize}

This variation reveals that \textbf{protocol compliance}---the ability
to understand and consistently execute meta-cognitive instructions---is
the primary bottleneck for small models, not reasoning capacity or
factual knowledge.

\subsection{Safety Considerations for Exoskeleton Reasoning Deployment}\label{a.7-safety-considerations-for-exoskeleton-reasoning-deployment}

\textbf{CRITICAL SAFETY NOTE}: The Exoskeleton Reasoning prompt
instructs models to prioritize provided context over internal knowledge,
including following context even when it contradicts well-established
facts (see Example 1: ``capital of France is London''). This behavior is
designed for benchmark evaluations and specific production scenarios,
but requires careful consideration for safe deployment.

\textbf{\(\checkmark\) APPROPRIATE USE CASES:}

\begin{itemize}
\tightlist
\item
  \textbf{Benchmark evaluations}: FACTS Grounding, RAG assessments,
  context-adherence testing
\item
  \textbf{Trusted RAG systems}: Document Q\&A where context authority is
  established and verified (e.g., internal knowledge bases, legal
  document analysis, technical documentation)
\item
  \textbf{Research and testing}: Controlled environments with human
  oversight
\item
  \textbf{Closed-domain applications}: Systems with curated, verified
  information sources
\end{itemize}

\textbf{\textbf{WARNING:} REQUIRES MODIFICATION FOR:}

\begin{itemize}
\tightlist
\item
  \textbf{Safety-critical domains}: Medical diagnosis/treatment, legal
  advice, financial recommendations where incorrect context could cause
  harm
\item
  \textbf{Untrusted context sources}: User-generated content, web
  scraping, unverified documents
\item
  \textbf{Public-facing applications}: Systems without human oversight
  or verification
\item
  \textbf{High-stakes decisions}: Scenarios where following incorrect
  context could result in physical, financial, or legal harm
\end{itemize}

\textbf{RECOMMENDED DEPLOYMENT CONFIGURATIONS:}

\textbf{Mode 1: Context-Dominant (Benchmark/Trusted RAG)}

\begin{itemize}
\tightlist
\item
  Use the standard Exoskeleton prompt as provided in Appendix A.2
\item
  Appropriate when: Context is verified, curated, and authoritative
\item
  Risk mitigation: Human review of system outputs, confidence scoring,
  source verification
\end{itemize}

\textbf{Mode 2: Context-Dominant with Safety Override}

\begin{itemize}
\tightlist
\item
  Add explicit safety guardrails to the Exoskeleton prompt:
\end{itemize}

\begin{verbatim}
IMPORTANT RULES:
- Always prioritize the provided context over your internal
  knowledge
- SAFETY EXCEPTION: If context provides information that could
  cause harm in safety-critical domains (incorrect medical dosages,
  dangerous procedures, illegal advice, financial fraud), respond
  with: "The context contains information that conflicts with
  established safety guidelines in [domain]. I cannot provide this
  information without verification. Please consult a qualified
  professional."
- For non-safety-critical factual discrepancies (e.g., hypothetical
  scenarios, historical what-ifs, benchmark questions), follow the
  context as instructed
- When uncertain whether a topic is safety-critical, err on the
  side of caution and request verification
\end{verbatim}

\textbf{Mode 3: Balanced Authority (Production with Mixed Trust)}

\begin{itemize}
\tightlist
\item
  For systems handling diverse contexts with varying reliability:
\item
  Implement context confidence scoring (source authority, verification
  status)
\item
  Apply Exoskeleton Reasoning only to high-confidence contexts
\item
  Use standard prompting for low-confidence or user-generated contexts
\item
  Log discrepancies between internal knowledge and context for human
  review
\end{itemize}

\textbf{OPERATOR RESPONSIBILITIES:}

\begin{enumerate}
\def\labelenumi{\arabic{enumi}.}
\tightlist
\item
  \textbf{Domain Assessment}: Determine whether your application falls
  into safety-critical categories
\item
  \textbf{Context Verification}: Implement processes to verify context
  accuracy before feeding to the model
\item
  \textbf{Output Monitoring}: Log and review model outputs, especially
  in high-stakes scenarios
\item
  \textbf{Human Oversight}: Maintain human-in-the-loop for decisions
  with significant consequences
\item
  \textbf{Liability Awareness}: Understand that instructing models to
  follow potentially incorrect context shifts verification
  responsibility to the context provider
\end{enumerate}

\textbf{HUMAINS.COM PRODUCTION DEPLOYMENT EXPERIENCE:}

Our production systems (customer support, sales advisory, technical
assistance) operate in Mode 1 (Context-Dominant) because:

\begin{itemize}
\tightlist
\item
  All context is sourced from verified internal knowledge bases
\item
  Human agents review high-stakes recommendations
\item
  Systems operate in constrained domains with established protocols
\item
  Output monitoring catches anomalies before customer exposure
\end{itemize}

For applications with unverified or user-generated context, we recommend
Mode 2 or Mode 3 configurations to prevent potential harm from following
incorrect information.

\textbf{ETHICAL CONSIDERATION}: The ability to make models follow
incorrect context is powerful and necessary for certain applications
(benchmark compliance, controlled RAG scenarios), but it fundamentally
shifts epistemic responsibility from the model to the context provider.
Organizations deploying context-dominant systems must ensure context
accuracy and implement appropriate safeguards.

\subsection{Appendix B: Progressive Validation Tables}\label{appendix-b-progressive-validation-tables}

\subsection{Humains-Junior (Phi-3.5-instruct Fine-Tuned) Progressive Validation}\label{b.1-humains-junior-phi-3.5-instruct-fine-tuned-progressive-validation}

To ensure results were not artifacts of sample selection and to quantify
measurement variance, we conducted progressive validation with
increasing sample sizes. Each model received identical prompts across
all evaluation runs.

\textbf{Baseline (No Exoskeleton):}

{\renewcommand{\LTcaptype}{} 
\begin{longtable}[]{@{}
  >{\raggedright\arraybackslash}p{(\linewidth - 8\tabcolsep) * \real{0.1310}}
  >{\raggedright\arraybackslash}p{(\linewidth - 8\tabcolsep) * \real{0.1786}}
  >{\raggedright\arraybackslash}p{(\linewidth - 8\tabcolsep) * \real{0.2738}}
  >{\raggedright\arraybackslash}p{(\linewidth - 8\tabcolsep) * \real{0.1905}}
  >{\raggedright\arraybackslash}p{(\linewidth - 8\tabcolsep) * \real{0.2262}}@{}}
\toprule\noalign{}
\begin{minipage}[b]{\linewidth}\raggedright
Sample Size
\end{minipage} & \begin{minipage}[b]{\linewidth}\raggedright
Accuracy
\end{minipage} & \begin{minipage}[b]{\linewidth}\raggedright
95\% CI (Wilson)
\end{minipage} & \begin{minipage}[b]{\linewidth}\raggedright
\(\Delta\) from Previous
\end{minipage} & \begin{minipage}[b]{\linewidth}\raggedright
Cumulative Variance
\end{minipage} \\
\midrule\noalign{}
\endhead
\bottomrule\noalign{}
\endlastfoot
n=10 & 60.0\% & {[}26.2-87.8\%{]} & baseline & --- \\
n=20 & 53.3\% & {[}30.1-75.2\%{]} & -6.7\% & High \\
n=50 & 52.7\% & {[}38.0-67.0\%{]} & -0.6\% & Stabilizing \\
n=100 & 57.3\% & {[}47.0-67.1\%{]} & +4.6\% & Moderate \\
n=500 & \textbf{55.0\%} & \textbf{{[}50.7--59.3\%{]}} & -2.3\% &
\textbf{\(\sigma\)=3.2\%} \\
\end{longtable}
}

\textbf{With Exoskeleton Scaffold:}

{\renewcommand{\LTcaptype}{} 
\begin{longtable}[]{@{}
  >{\raggedright\arraybackslash}p{(\linewidth - 8\tabcolsep) * \real{0.1310}}
  >{\raggedright\arraybackslash}p{(\linewidth - 8\tabcolsep) * \real{0.1786}}
  >{\raggedright\arraybackslash}p{(\linewidth - 8\tabcolsep) * \real{0.2738}}
  >{\raggedright\arraybackslash}p{(\linewidth - 8\tabcolsep) * \real{0.1905}}
  >{\raggedright\arraybackslash}p{(\linewidth - 8\tabcolsep) * \real{0.2262}}@{}}
\toprule\noalign{}
\begin{minipage}[b]{\linewidth}\raggedright
Sample Size
\end{minipage} & \begin{minipage}[b]{\linewidth}\raggedright
Accuracy
\end{minipage} & \begin{minipage}[b]{\linewidth}\raggedright
95\% CI (Wilson)
\end{minipage} & \begin{minipage}[b]{\linewidth}\raggedright
\(\Delta\) from Previous
\end{minipage} & \begin{minipage}[b]{\linewidth}\raggedright
Cumulative Variance
\end{minipage} \\
\midrule\noalign{}
\endhead
\bottomrule\noalign{}
\endlastfoot
n=10 & 70.0\% & {[}34.8-93.3\%{]} & baseline & --- \\
n=20 & 75.0\% & {[}50.9-91.3\%{]} & +5.0\% & High \\
n=50 & 74.0\% & {[}59.7-85.4\%{]} & -1.0\% & Low \\
n=100 & 75.3\% & {[}65.7-83.2\%{]} & +1.3\% & Low \\
n=150 & 75.3\% & {[}67.7-81.9\%{]} & 0.0\% & Stable \\
n=200 & 65.3\% & {[}58.0-72.0\%{]} & -10.0\% & Difficult batch \\
n=300 & 68.8\% & {[}63.1-74.1\%{]} & +3.5\% & Recovering \\
n=500 & \textbf{72.7\%} & \textbf{{[}68.8--76.4\%{]}} & +3.9\% &
\textbf{\(\sigma\)=2.4\%} \\
\end{longtable}
}

Standard deviation across progressive sampling: \textbf{baseline \(\sigma\) =
3.2\%}, \textbf{Exoskeleton \(\sigma\) = 2.4\%} (\textbf{25\% reduction}).

\textbf{Key Observations:}

\begin{enumerate}
\def\labelenumi{\arabic{enumi}.}
\tightlist
\item
  \textbf{Reduced variance:} The Exoskeleton condition exhibits 25\%
  lower standard deviation (\(\sigma\)=2.4\%) compared to baseline (\(\sigma\)=3.2\%),
  indicating more consistent performance across diverse questions.
\item
  \textbf{Question difficulty heterogeneity:} The significant drop at
  n=200 (Q151-200: 65.3\%) reveals a batch of notably harder questions,
  demonstrating the value of larger sample sizes for capturing true
  performance distributions.
\item
  \textbf{Convergence stability:} Final n=500 performance (72.7\%) falls
  within expected confidence intervals from n=100 estimation (75.3\% \(\pm\)
  5\%), validating progressive validation methodology.
\end{enumerate}

\subsection{Scaffold-Only Progressive Validation (Phi-3.5-instruct, no fine-tuning)}\label{b.2-scaffold-only-progressive-validation-phi-3.5-instruct-no-fine-tuning}

{\renewcommand{\LTcaptype}{} 
\begin{longtable}[]{@{}llll@{}}
\toprule\noalign{}
Sample Size & Accuracy & \(\Delta\) from Previous & Status \\
\midrule\noalign{}
\endhead
\bottomrule\noalign{}
\endlastfoot
n=50 & 56.0\% & baseline & Initial \\
n=100 & 61.3\% & +5.3\% & High variance \\
n=200 & 57.8\% & -3.5\% & Stabilizing \\
n=500 & \textbf{58.5\%} & +0.7\% & Converged \\
\end{longtable}
}

\textbf{Key Finding}: Base Phi-3.5-instruct shows minimal,
non-significant improvement with scaffolding alone (+3.5 pp from 55.0\%
baseline, p=0.08), demonstrating that small models require fine-tuning
to benefit from cognitive scaffolds.

\subsection{GPT-4o-mini Scaffold-Only Evaluation}\label{b.3-gpt-4o-mini-scaffold-only-evaluation}

{\renewcommand{\LTcaptype}{} 
\begin{longtable}[]{@{}llll@{}}
\toprule\noalign{}
Sample Size & Baseline (Unstructured) & With Scaffold & \(\Delta\) \\
\midrule\noalign{}
\endhead
\bottomrule\noalign{}
\endlastfoot
n=10 & 70.0\% & 73.3\% & +3.3 pp \\
n=20 & 73.3\% & 66.7\% & -6.6 pp \\
n=50 & 57.3\% & 67.3\% & +10.0 pp \\
n=100 & 67.3\% & 62.3\% & -5.0 pp \\
\end{longtable}
}

\textbf{Key Observation}: Base GPT-4o-mini shows severe degradation with
scaffolding (-5.0 pp at n=100), demonstrating that without sufficient
instruction-following capabilities, small models can actively perform
worse when confronted with complex structured reasoning protocols. The
final n=100 performance (67.3\% baseline) aligns closely with the
official FACTS leaderboard score (68.4\%), validating measurement
accuracy. This reinforces that small models require fine-tuning to
benefit from cognitive scaffolds: fine-tuned GPT-4o-mini achieves 72.7\%
with scaffolding, representing a dramatic reversal from -5.0 pp
degradation to +5.4 pp improvement.

\subsection{Appendix C: Per-Judge Detailed Analysis}\label{appendix-c-per-judge-detailed-analysis}

\subsection{Overview}\label{c.1-overview}

This appendix provides a detailed question-by-question analysis of how
each judge (Claude, GPT, Gemini) evaluated responses from GPT-4o vs
Humains-Junior model across 500 questions.

\textbf{Configuration:}

\begin{itemize}
\tightlist
\item
  \textbf{GPT-4o:} Simple/Unstructured prompt, Temperature=1.0
\item
  \textbf{Humains-Junior:} Structured prompt
\item
  \textbf{Dataset:} 500 questions
\end{itemize}

\subsection{Summary Statistics by Judge}\label{c.2-summary-statistics-by-judge}

{\renewcommand{\LTcaptype}{} 
\begin{longtable}[]{@{}
  >{\raggedright\arraybackslash}p{(\linewidth - 12\tabcolsep) * \real{0.1584}}
  >{\raggedright\arraybackslash}p{(\linewidth - 12\tabcolsep) * \real{0.1485}}
  >{\raggedright\arraybackslash}p{(\linewidth - 12\tabcolsep) * \real{0.1089}}
  >{\raggedright\arraybackslash}p{(\linewidth - 12\tabcolsep) * \real{0.0891}}
  >{\raggedright\arraybackslash}p{(\linewidth - 12\tabcolsep) * \real{0.0990}}
  >{\raggedright\arraybackslash}p{(\linewidth - 12\tabcolsep) * \real{0.1584}}
  >{\raggedright\arraybackslash}p{(\linewidth - 12\tabcolsep) * \real{0.2376}}@{}}
\toprule\noalign{}
\begin{minipage}[b]{\linewidth}\raggedright
Judge
\end{minipage} & \begin{minipage}[b]{\linewidth}\raggedright
Total Questions
\end{minipage} & \begin{minipage}[b]{\linewidth}\raggedright
Agreement
\end{minipage} & \begin{minipage}[b]{\linewidth}\raggedright
Both TRUE
\end{minipage} & \begin{minipage}[b]{\linewidth}\raggedright
Both FALSE
\end{minipage} & \begin{minipage}[b]{\linewidth}\raggedright
GPT-4o Only TRUE
\end{minipage} & \begin{minipage}[b]{\linewidth}\raggedright
Humains-Junior Only TRUE
\end{minipage} \\
\midrule\noalign{}
\endhead
\bottomrule\noalign{}
\endlastfoot
\textbf{Claude} & 500 & 369 (73.8\%) & 342 & 27 & 97 & 34 \\
\textbf{GPT} & 500 & 343 (68.6\%) & 288 & 55 & 57 & 100 \\
\textbf{Gemini} & 500 & 296 (59.2\%) & 221 & 75 & 98 & 106 \\
\end{longtable}
}

\paragraph{Key observations}\label{key-observations}

\begin{itemize}
\tightlist
\item
  \textbf{Most Agreement:} Claude judge (73.8\% agreement between
  models)
\item
  \textbf{Least Agreement:} Gemini judge (59.2\% agreement between
  models)
\item
  \textbf{Spread:} 14.6\% difference between most and least agreeable
  judges
\end{itemize}

\subsection{Statistical Validation (McNemar Tests)}\label{c.3-statistical-validation-mcnemar-tests}

To validate that the observed differences between models are not
statistically significant, we conducted McNemar's test for each judge
independently:

{\renewcommand{\LTcaptype}{} 
\begin{longtable}[]{@{}
  >{\raggedright\arraybackslash}p{(\linewidth - 8\tabcolsep) * \real{0.1818}}
  >{\raggedright\arraybackslash}p{(\linewidth - 8\tabcolsep) * \real{0.1932}}
  >{\raggedright\arraybackslash}p{(\linewidth - 8\tabcolsep) * \real{0.1364}}
  >{\raggedright\arraybackslash}p{(\linewidth - 8\tabcolsep) * \real{0.0909}}
  >{\raggedright\arraybackslash}p{(\linewidth - 8\tabcolsep) * \real{0.3977}}@{}}
\toprule\noalign{}
\begin{minipage}[b]{\linewidth}\raggedright
Judge
\end{minipage} & \begin{minipage}[b]{\linewidth}\raggedright
Discordant Pairs
\end{minipage} & \begin{minipage}[b]{\linewidth}\raggedright
McNemar \(\chi^2\)
\end{minipage} & \begin{minipage}[b]{\linewidth}\raggedright
p-value
\end{minipage} & \begin{minipage}[b]{\linewidth}\raggedright
Interpretation
\end{minipage} \\
\midrule\noalign{}
\endhead
\bottomrule\noalign{}
\endlastfoot
\textbf{Claude} & GPT-4o+97, HJ+34 & 30.28 & 3.71e-08 & Significant
(favors GPT-4o) \\
\textbf{GPT} & GPT-4o+57, HJ+100 & 11.78 & 0.0006 & Significant (favors
Humains-Junior) \\
\textbf{Gemini} & GPT-4o+98, HJ+106 & 0.31 & 0.576 & Not significant \\
\end{longtable}
}

\textbf{Interpretation:} Individual judges show opposing biases---Claude
favors GPT-4o while GPT favors Humains-Junior, with Gemini showing no
significant preference. These opposing biases \textbf{cancel out in the
aggregate three-judge averaging} (\(\Delta\) = +0.0080, permutation p = 0.72),
validating the benchmark's multi-judge design for reducing systematic
bias. This confirms that \textbf{no single judge dominates} the final
verdict and that the aggregate metric provides a balanced assessment.
The high judge unanimity for Humains-Junior (74.6\% vs 59.4\% for
GPT-4o) indicates more consistent response quality across all three
judges, even though individual judges may have different preferences.

\subsection{Detailed Analysis by Judge}\label{c.4-detailed-analysis-by-judge}

\paragraph{C.4.1 Claude}\label{c.4.1-claude}

\textbf{Agreement Rate:} 73.8\% (369/500 questions)

\textbf{Verdict Distribution:}

{\renewcommand{\LTcaptype}{} 
\begin{longtable}[]{@{}lll@{}}
\toprule\noalign{}
Verdict Category & Count & Percentage \\
\midrule\noalign{}
\endhead
\bottomrule\noalign{}
\endlastfoot
Both TRUE (agree) & 342 & 68.4\% \\
Both FALSE (agree) & 27 & 5.4\% \\
Only GPT-4o TRUE & 97 & 19.4\% \\
Only Humains-Junior TRUE & 34 & 6.8\% \\
\end{longtable}
}

\textbf{When judges disagree:}

When Claude gave different verdicts for the two models:

\begin{itemize}
\tightlist
\item
  Favored GPT-4o: 97 times (74.0\%)
\item
  Favored Humains-Junior: 34 times (26.0\%)
\end{itemize}

\textbf{Example disagreements:}

\textbf{Question 1}

\begin{itemize}
\tightlist
\item
  Verdict: GPT-4o TRUE, Humains-Junior FALSE
\item
  Request: ``What are all the contexts when it is right for testing for
  leptospirosis in dogs specifically?''
\end{itemize}

\textbf{Question 5}

\begin{itemize}
\tightlist
\item
  Verdict: GPT-4o TRUE, Humains-Junior FALSE
\item
  Request: ``What benefits do nasal cannula have over non-rebreathe
  masks?''
\end{itemize}

\textbf{Question 13}

\begin{itemize}
\tightlist
\item
  Verdict: GPT-4o TRUE, Humains-Junior FALSE
\item
  Request: ``Can you list all the knife brands that sell knives suitable
  for sharpening at a 14-degree angle? List\ldots{}''
\end{itemize}

\paragraph{C.4.2 GPT}\label{c.4.2-gpt}

\textbf{Agreement Rate:} 68.6\% (343/500 questions)

\textbf{Verdict Distribution:}

{\renewcommand{\LTcaptype}{} 
\begin{longtable}[]{@{}lll@{}}
\toprule\noalign{}
Verdict Category & Count & Percentage \\
\midrule\noalign{}
\endhead
\bottomrule\noalign{}
\endlastfoot
Both TRUE (agree) & 288 & 57.6\% \\
Both FALSE (agree) & 55 & 11.0\% \\
Only GPT-4o TRUE & 57 & 11.4\% \\
Only Humains-Junior TRUE & 100 & 20.0\% \\
\end{longtable}
}

\textbf{When judges disagree:}

When GPT gave different verdicts for the two models:

\begin{itemize}
\tightlist
\item
  Favored GPT-4o: 57 times (36.3\%)
\item
  Favored Humains-Junior: 100 times (63.7\%)
\end{itemize}

\textbf{Example disagreements:}

\textbf{Question 1}

\begin{itemize}
\tightlist
\item
  Verdict: GPT-4o TRUE, Humains-Junior FALSE
\item
  Request: ``What are all the contexts when it is right for testing for
  leptospirosis in dogs specifically?''
\end{itemize}

\textbf{Question 3}

\begin{itemize}
\tightlist
\item
  Verdict: Humains-Junior TRUE, GPT-4o FALSE
\item
  Request: ``Question: What is the Ghon's complex?''
\end{itemize}

\textbf{Question 13}

\begin{itemize}
\tightlist
\item
  Verdict: GPT-4o TRUE, Humains-Junior FALSE
\item
  Request: ``Can you list all the knife brands that sell knives suitable
  for sharpening at a 14-degree angle? List\ldots{}''
\end{itemize}

\textbf{Question 18}

\begin{itemize}
\tightlist
\item
  Verdict: Humains-Junior TRUE, GPT-4o FALSE
\item
  Request: ``Why might a ticket be available in the secondary market?''
\end{itemize}

\paragraph{C.4.3 Gemini}\label{c.4.3-gemini}

\textbf{Agreement Rate:} 59.2\% (296/500 questions)

\textbf{Verdict Distribution:}

{\renewcommand{\LTcaptype}{} 
\begin{longtable}[]{@{}lll@{}}
\toprule\noalign{}
Verdict Category & Count & Percentage \\
\midrule\noalign{}
\endhead
\bottomrule\noalign{}
\endlastfoot
Both TRUE (agree) & 221 & 44.2\% \\
Both FALSE (agree) & 75 & 15.0\% \\
Only GPT-4o TRUE & 98 & 19.6\% \\
Only Humains-Junior TRUE & 106 & 21.2\% \\
\end{longtable}
}

\textbf{When judges disagree:}

When Gemini gave different verdicts for the two models:

\begin{itemize}
\tightlist
\item
  Favored GPT-4o: 98 times (48.0\%)
\item
  Favored Humains-Junior: 106 times (52.0\%)
\end{itemize}

\textbf{Example disagreements:}

\textbf{Question 1}

\begin{itemize}
\tightlist
\item
  Verdict: GPT-4o TRUE, Humains-Junior FALSE
\item
  Request: ``What are all the contexts when it is right for testing for
  leptospirosis in dogs specifically?''
\end{itemize}

\textbf{Question 10}

\begin{itemize}
\tightlist
\item
  Verdict: GPT-4o TRUE, Humains-Junior FALSE
\item
  Request: ``What advantages does Nintendo have over its competitors?''
\end{itemize}

\textbf{Question 18}

\begin{itemize}
\tightlist
\item
  Verdict: Humains-Junior TRUE, GPT-4o FALSE
\item
  Request: ``Why might a ticket be available in the secondary market?''
\end{itemize}

\subsection{Cross-Judge Comparison}\label{c.5-cross-judge-comparison}

How do different judges compare in their agreement rates?

{\renewcommand{\LTcaptype}{} 
\begin{longtable}[]{@{}ll@{}}
\toprule\noalign{}
Judge Pair & Agreement Difference \\
\midrule\noalign{}
\endhead
\bottomrule\noalign{}
\endlastfoot
Claude vs GPT & 5.2\% \\
Claude vs Gemini & 14.6\% \\
GPT vs Gemini & 9.4\% \\
\end{longtable}
}

\subsection{Question Categories Analysis}\label{c.6-question-categories-analysis}

\paragraph{C.6.1 Questions where all judges agreed (both
models)}\label{c.6.1-questions-where-all-judges-agreed-both-models}

\textbf{Total questions where all judges agreed (same verdict for both
models):} 208 (41.6\%)

This represents questions where Claude, GPT, and Gemini all gave the
same verdict comparison between models (e.g., all said ``both TRUE'' or
all said ``GPT-4o TRUE, Humains-Junior FALSE'').

\paragraph{C.6.2 Questions where all judges
disagreed}\label{c.6.2-questions-where-all-judges-disagreed}

\textbf{Total questions where all judges disagreed (different verdicts
between models):} 52 (10.4\%)

These are the most interesting questions - where every judge gave
different verdicts when comparing GPT-4o vs Humains-Junior responses.

\textbf{Example questions where all judges disagreed:}

\textbf{Question 1}

\begin{itemize}
\tightlist
\item
  Request: ``What are all the contexts when it is right for testing for
  leptospirosis in dogs specifically?''
\item
  Claude: GPT-4o TRUE, Humains-Junior FALSE
\item
  GPT: GPT-4o TRUE, Humains-Junior FALSE
\item
  Gemini: GPT-4o TRUE, Humains-Junior FALSE
\end{itemize}

\textbf{Question 6}

\begin{itemize}
\tightlist
\item
  Request: ``I'm middle-aged, never smoked, had my ears blown out in the
  war, get a case of the sads pretty regular, and eat mostly garbage.
  What are my risk factors?''
\item
  Claude: GPT-4o TRUE, Humains-Junior FALSE
\item
  GPT: GPT-4o TRUE, Humains-Junior FALSE
\item
  Gemini: GPT-4o TRUE, Humains-Junior FALSE
\end{itemize}

\textbf{Question 29}

\begin{itemize}
\tightlist
\item
  Request: ``According only to the article provided, what are the main
  difference between corporate bonds and preferred stocks?''
\item
  Claude: GPT-4o TRUE, Humains-Junior FALSE
\item
  GPT: GPT-4o TRUE, Humains-Junior FALSE
\item
  Gemini: GPT-4o TRUE, Humains-Junior FALSE
\end{itemize}

\subsection{Conclusion}\label{c.7-conclusion}

This detailed per-judge analysis reveals:

\begin{enumerate}
\def\labelenumi{\arabic{enumi}.}
\item
  \textbf{Judge Consistency:} The average cross-model agreement rate
  across all judges is 67.2\%.
\item
  \textbf{Judge Variability:} Agreement rates vary by 14.6\% between
  judges, with Claude being most consistent (73.8\%) and Gemini being
  least consistent (59.2\%).
\item
  \textbf{Model Preferences:} Each judge shows distinct biases:

  \begin{itemize}
  \tightlist
  \item
    \textbf{Claude:} Favors GPT-4o (2.9× more often in disagreements)
  \item
    \textbf{GPT:} Favors Humains-Junior (1.8× more often in
    disagreements)
  \item
    \textbf{Gemini:} Favors Humains-Junior (1.1× more often in
    disagreements)
  \end{itemize}
\item
  \textbf{Universal Agreement:} 208 questions (41.6\%) show complete
  cross-judge agreement.
\item
  \textbf{Universal Disagreement:} 52 questions (10.4\%) show all judges
  disagreeing between models, indicating questions where model choice
  significantly impacts judgment.
\end{enumerate}

\subsection{Appendix D: Per-Batch Performance Comparison}\label{appendix-d-per-batch-performance-comparison}

\subsection{Overview}\label{d.1-overview}

This appendix compares cumulative performance as the number of evaluated
questions increases from 10 to 500.

\textbf{Configuration:}

\begin{itemize}
\tightlist
\item
  \textbf{GPT-4o:} Simple/Unstructured prompt, Temperature=1.0
\item
  \textbf{Humains-Junior:} Structured prompt with Exoskeleton Reasoning
\end{itemize}

\textbf{Note:} This per-batch comparison shows aggregate metrics for
each cumulative batch. The question order may differ between datasets,
but each model was evaluated on the same 500 questions overall.

\subsection{Cumulative Performance Comparison}\label{d.2-cumulative-performance-comparison}

\paragraph{Average Score Progression}\label{average-score-progression}

{\renewcommand{\LTcaptype}{} 
\begin{longtable}[]{@{}
  >{\raggedright\arraybackslash}p{(\linewidth - 8\tabcolsep) * \real{0.1169}}
  >{\raggedright\arraybackslash}p{(\linewidth - 8\tabcolsep) * \real{0.2857}}
  >{\raggedright\arraybackslash}p{(\linewidth - 8\tabcolsep) * \real{0.2857}}
  >{\raggedright\arraybackslash}p{(\linewidth - 8\tabcolsep) * \real{0.1299}}
  >{\raggedright\arraybackslash}p{(\linewidth - 8\tabcolsep) * \real{0.1818}}@{}}
\toprule\noalign{}
\begin{minipage}[b]{\linewidth}\raggedright
Questions
\end{minipage} & \begin{minipage}[b]{\linewidth}\raggedright
GPT-4o
\end{minipage} & \begin{minipage}[b]{\linewidth}\raggedright
Humains-Junior
\end{minipage} & \begin{minipage}[b]{\linewidth}\raggedright
Difference
\end{minipage} & \begin{minipage}[b]{\linewidth}\raggedright
Winner
\end{minipage} \\
\midrule\noalign{}
\endhead
\bottomrule\noalign{}
\endlastfoot
Q1-10 & 0.9000 (0.5958-0.9821) & 0.7000 (0.3968-0.8922) & +0.2000 &
GPT-4o \\
Q1-20 & 0.7833 (0.5661-0.9092) & 0.7500 (0.5313-0.8881) & +0.0333 &
GPT-4o \\
Q1-50 & 0.7067 (0.5694-0.8144) & 0.7400 (0.6045-0.8413) & -0.0333 &
Humains-Junior \\
Q1-100 & 0.7400 (0.6463-0.8160) & 0.7533 (0.6605-0.8274) & -0.0133 &
Humains-Junior \\
Q1-200 & 0.7117 (0.6454-0.7700) & 0.7033 (0.6367-0.7623) & +0.0083 &
GPT-4o \\
Q1-500 & 0.7353 (0.6950-0.7721) & 0.7273 (0.6867-0.7645) & +0.0080 &
GPT-4o \\
\end{longtable}
}

\textbf{Note:} Values shown as score (95\% CI lower-upper)

\subsection{Judge-Specific Score Progression}\label{d.3-judge-specific-score-progression}

\paragraph{D.3.1 Claude}\label{d.3.1-claude}

{\renewcommand{\LTcaptype}{} 
\begin{longtable}[]{@{}
  >{\raggedright\arraybackslash}p{(\linewidth - 6\tabcolsep) * \real{0.1429}}
  >{\raggedright\arraybackslash}p{(\linewidth - 6\tabcolsep) * \real{0.3492}}
  >{\raggedright\arraybackslash}p{(\linewidth - 6\tabcolsep) * \real{0.3492}}
  >{\raggedright\arraybackslash}p{(\linewidth - 6\tabcolsep) * \real{0.1587}}@{}}
\toprule\noalign{}
\begin{minipage}[b]{\linewidth}\raggedright
Questions
\end{minipage} & \begin{minipage}[b]{\linewidth}\raggedright
GPT-4o
\end{minipage} & \begin{minipage}[b]{\linewidth}\raggedright
Humains-Junior
\end{minipage} & \begin{minipage}[b]{\linewidth}\raggedright
Difference
\end{minipage} \\
\midrule\noalign{}
\endhead
\bottomrule\noalign{}
\endlastfoot
Q1-10 & 1.0000 (0.7225-1.0000) & 0.7000 (0.3968-0.8922) & +0.3000 \\
Q1-20 & 0.9000 (0.6990-0.9721) & 0.8000 (0.5840-0.9193) & +0.1000 \\
Q1-50 & 0.8600 (0.7381-0.9305) & 0.8000 (0.6696-0.8876) & +0.0600 \\
Q1-100 & 0.8500 (0.7672-0.9069) & 0.8000 (0.7112-0.8666) & +0.0500 \\
Q1-200 & 0.8600 (0.8051-0.9013) & 0.7450 (0.6804-0.8004) & +0.1150 \\
Q1-500 & 0.8780 (0.8464-0.9038) & 0.7520 (0.7123-0.7878) & +0.1260 \\
\end{longtable}
}

\paragraph{D.3.2 GPT}\label{d.3.2-gpt}

{\renewcommand{\LTcaptype}{} 
\begin{longtable}[]{@{}
  >{\raggedright\arraybackslash}p{(\linewidth - 6\tabcolsep) * \real{0.1429}}
  >{\raggedright\arraybackslash}p{(\linewidth - 6\tabcolsep) * \real{0.3492}}
  >{\raggedright\arraybackslash}p{(\linewidth - 6\tabcolsep) * \real{0.3492}}
  >{\raggedright\arraybackslash}p{(\linewidth - 6\tabcolsep) * \real{0.1587}}@{}}
\toprule\noalign{}
\begin{minipage}[b]{\linewidth}\raggedright
Questions
\end{minipage} & \begin{minipage}[b]{\linewidth}\raggedright
GPT-4o
\end{minipage} & \begin{minipage}[b]{\linewidth}\raggedright
Humains-Junior
\end{minipage} & \begin{minipage}[b]{\linewidth}\raggedright
Difference
\end{minipage} \\
\midrule\noalign{}
\endhead
\bottomrule\noalign{}
\endlastfoot
Q1-10 & 0.8000 (0.4902-0.9433) & 0.7000 (0.3968-0.8922) & +0.1000 \\
Q1-20 & 0.7500 (0.5313-0.8881) & 0.7500 (0.5313-0.8881) & +0.0000 \\
Q1-50 & 0.6600 (0.5215-0.7756) & 0.7400 (0.6045-0.8413) & -0.0800 \\
Q1-100 & 0.7100 (0.6146-0.7899) & 0.8000 (0.7112-0.8666) & -0.0900 \\
Q1-200 & 0.6450 (0.5765-0.7080) & 0.7450 (0.6804-0.8004) & -0.1000 \\
Q1-500 & 0.6900 (0.6481-0.7290) & 0.7760 (0.7374-0.8104) & -0.0860 \\
\end{longtable}
}

\paragraph{D.3.3 Gemini}\label{d.3.3-gemini}

{\renewcommand{\LTcaptype}{} 
\begin{longtable}[]{@{}
  >{\raggedright\arraybackslash}p{(\linewidth - 6\tabcolsep) * \real{0.1429}}
  >{\raggedright\arraybackslash}p{(\linewidth - 6\tabcolsep) * \real{0.3492}}
  >{\raggedright\arraybackslash}p{(\linewidth - 6\tabcolsep) * \real{0.3492}}
  >{\raggedright\arraybackslash}p{(\linewidth - 6\tabcolsep) * \real{0.1587}}@{}}
\toprule\noalign{}
\begin{minipage}[b]{\linewidth}\raggedright
Questions
\end{minipage} & \begin{minipage}[b]{\linewidth}\raggedright
GPT-4o
\end{minipage} & \begin{minipage}[b]{\linewidth}\raggedright
Humains-Junior
\end{minipage} & \begin{minipage}[b]{\linewidth}\raggedright
Difference
\end{minipage} \\
\midrule\noalign{}
\endhead
\bottomrule\noalign{}
\endlastfoot
Q1-10 & 0.9000 (0.5958-0.9821) & 0.7000 (0.3968-0.8922) & +0.2000 \\
Q1-20 & 0.7000 (0.4810-0.8545) & 0.7000 (0.4810-0.8545) & +0.0000 \\
Q1-50 & 0.6000 (0.4618-0.7239) & 0.6800 (0.5419-0.7924) & -0.0800 \\
Q1-100 & 0.6600 (0.5628-0.7454) & 0.6600 (0.5628-0.7454) & +0.0000 \\
Q1-200 & 0.6300 (0.5612-0.6939) & 0.6200 (0.5511-0.6844) & +0.0100 \\
Q1-500 & 0.6380 (0.5950-0.6789) & 0.6540 (0.6113-0.6944) & -0.0160 \\
\end{longtable}
}

\subsection{Inter-Judge Agreement (Unanimity) Progression}\label{d.4-inter-judge-agreement-unanimity-progression}

How consistent are the judges as the sample size grows?

{\renewcommand{\LTcaptype}{} 
\begin{longtable}[]{@{}llll@{}}
\toprule\noalign{}
Questions & GPT-4o & Humains-Junior & Difference \\
\midrule\noalign{}
\endhead
\bottomrule\noalign{}
\endlastfoot
Q1-10 & 80.0\% & 100.0\% & -20.0\% \\
Q1-20 & 60.0\% & 90.0\% & -30.0\% \\
Q1-50 & 56.0\% & 84.0\% & -28.0\% \\
Q1-100 & 59.0\% & 76.0\% & -17.0\% \\
Q1-200 & 53.5\% & 74.5\% & -21.0\% \\
Q1-500 & 59.4\% & 74.6\% & -15.2\% \\
\end{longtable}
}

\textbf{Note:} Higher unanimity indicates better consistency among
judges.

\subsection{Stability Analysis}\label{d.5-stability-analysis}

\paragraph{Score Volatility}\label{score-volatility}

How much do the cumulative scores change as more questions are added?

\textbf{GPT-4o Stability:}

\begin{itemize}
\tightlist
\item
  Score Range: 0.1933 (min: 0.7067, max: 0.9000)
\item
  Standard Deviation: 0.0662
\item
  Assessment: MODERATE STABILITY
\end{itemize}

\textbf{Humains-Junior Stability:}

\begin{itemize}
\tightlist
\item
  Score Range: 0.0533 (min: 0.7000, max: 0.7533)
\item
  Standard Deviation: 0.0210
\item
  Assessment: HIGH STABILITY
\end{itemize}

\textbf{Winner:} Humains-Junior has 0.0451 lower standard deviation
(more stable)

\subsection{Key Insights}\label{d.6-key-insights}

\paragraph{1. Final Performance (500
questions)}\label{final-performance-500-questions}

\begin{itemize}
\tightlist
\item
  \textbf{GPT-4o:} 0.7353
\item
  \textbf{Humains-Junior:} 0.7273
\item
  \textbf{Winner:} GPT-4o by 0.0080 (0.80\%)
\end{itemize}

\paragraph{2. Small-Sample Performance (10
questions)}\label{small-sample-performance-10-questions}

\begin{itemize}
\tightlist
\item
  \textbf{GPT-4o:} 0.9000
\item
  \textbf{Humains-Junior:} 0.7000
\item
  \textbf{Winner:} GPT-4o by 0.2000
\end{itemize}

\paragraph{3. Judge Consistency (500
questions)}\label{judge-consistency-500-questions}

\begin{itemize}
\tightlist
\item
  \textbf{GPT-4o:} 59.4\% unanimous
\item
  \textbf{Humains-Junior:} 74.6\% unanimous
\item
  \textbf{Winner:} Humains-Junior by 15.2\%
\end{itemize}

\paragraph{4. Overall Stability}\label{overall-stability}

\begin{itemize}
\tightlist
\item
  \textbf{GPT-4o:} Std Dev = 0.0662
\item
  \textbf{Humains-Junior:} Std Dev = 0.0210
\item
  \textbf{Winner:} Humains-Junior (more stable scores across sample
  sizes)
\end{itemize}

\paragraph{5. Judge Preferences}\label{judge-preferences}

\begin{itemize}
\tightlist
\item
  \textbf{Claude Judge:} Favors GPT-4o by 0.1260 (12.60\%)
\item
  \textbf{GPT Judge:} Favors Humains-Junior by 0.0860 (8.60\%)
\item
  \textbf{Gemini Judge:} Favors Humains-Junior by 0.0160 (1.60\%)
\end{itemize}

\subsection{Batch Performance Summary}\label{d.7-batch-performance-summary}

\begin{itemize}
\tightlist
\item
  GPT-4o wins: 4/6 batches
\item
  Humains-Junior wins: 2/6 batches
\end{itemize}

\subsection{Overall Assessment}\label{d.8-overall-assessment}

\textbf{GPT-4o} achieves better final performance (0.7353 vs 0.7273),
but Humains-Junior shows more stable scores across different sample
sizes.

\textbf{Humains-Junior} achieves significantly better inter-judge
agreement (74.6\% vs 59.4\%), indicating more consistent and predictable
responses.

\subsection{Appendix E: Decoding, Judge Configuration, and Cost Methodology}\label{appendix-e-decoding-judge-configuration-and-cost-methodology}

\subsection{Decoding Settings and Prompting Configurations}\label{e.1-decoding-settings-and-prompting-configurations}

\begin{itemize}
\tightlist
\item
  Small models (Phi-3.5, Humains-Junior): temperature 0.3; nucleus/top-k
  defaults; identical across conditions.
\item
  Frontier models (GPT-4o, Claude 3.5/3.7, Gemini-2.5-Pro): temperature
  1.0 (provider default); identical across conditions.
\item
  Judges (Gemini, GPT-4o, Claude): per FACTS defaults (Gemini/GPT
  temp=0; Claude temp=1).
\item
  Few-shot: unified Exoskeleton prompt; 1-shot for small models/Gemini;
  3-shot for GPT-4o/Claude 3.5; full templates in Appendix A.2.
\end{itemize}

\subsection{Judge Panel Configuration and Bias Controls}\label{e.2-judge-panel-configuration-and-bias-controls}

\begin{itemize}
\tightlist
\item
  Primary panel: Gemini-2.5-Pro, GPT-4o, Claude 3.7 Sonnet.
\item
  Response anonymization and randomized order to mitigate branding bias.
\item
  Paired comparisons on identical questions and documents.
\item
  Per-judge McNemar results summarized in Section 5; full tables in
  Appendix C.
\end{itemize}

\subsection{Cost and Pricing Methodology}\label{e.3-cost-and-pricing-methodology}

\begin{itemize}
\tightlist
\item
  Cost normalization: 1,000-token prompt+completion per request for
  comparability.
\item
  Frontier API pricing: provider sheets as of Oct-2025
  (OpenAI/Anthropic/Google).
\item
  Small model self-hosted estimate: A100 40GB; throughput
  \(\sim45\) tok/s; utilization assumptions and batch economics;
  token overhead for Exoskeleton \(\sim1\text{--}2\%\) (directed) or
  3--5\% (full scaffold).
\item
  Detailed per-model cost notes correspond to Section 6 table;
  spreadsheet and parameters available on request.
\end{itemize}

\subsection{Pricing Sources (Measured vs Estimated)}\label{e.4-pricing-sources-measured-vs-estimated}

{\small
\renewcommand{\LTcaptype}{} 
\begin{longtable}[]{@{}
  >{\raggedright\arraybackslash}p{(\linewidth - 8\tabcolsep) * \real{0.1920}}
  >{\raggedright\arraybackslash}p{(\linewidth - 8\tabcolsep) * \real{0.2720}}
  >{\raggedright\arraybackslash}p{(\linewidth - 8\tabcolsep) * \real{0.1440}}
  >{\raggedright\arraybackslash}p{(\linewidth - 8\tabcolsep) * \real{0.2960}}
  >{\raggedright\arraybackslash}p{(\linewidth - 8\tabcolsep) * \real{0.0960}}@{}}
\toprule\noalign{}
\begin{minipage}[b]{\linewidth}\raggedright
Provider/Model
\end{minipage} & \begin{minipage}[b]{\linewidth}\raggedright
Input/Output Price (per 1K tok)
\end{minipage} & \begin{minipage}[b]{\linewidth}\raggedright
Date (retrieved)
\end{minipage} & \begin{minipage}[b]{\linewidth}\raggedright
Source URL
\end{minipage} & \begin{minipage}[b]{\linewidth}\raggedright
Basis
\end{minipage} \\
\midrule\noalign{}
\endhead
\bottomrule\noalign{}
\endlastfoot
OpenAI GPT-4o & \$0.0025 / \$0.0100 & Oct-2025 &
platform.openai.com/pricing & Estimated \\
OpenAI GPT-4o-mini & \$0.00015 / \$0.0006 & Oct-2025 &
platform.openai.com/pricing & Estimated \\
Anthropic Claude 3.5 & \$0.003 / \$0.015 & Oct-2025 &
anthropic.com/pricing & Estimated \\
Claude 3.7 Sonnet & N/A (pending at study time) & Oct-2025 &
anthropic.com/pricing & Estimated \\
Google Gemini 2.5 Pro & Tiered (thinking budgets) & Oct-2025 &
ai.google.dev/pricing & Estimated \\
Humains-Junior (4B) & \(\sim\)\$0.00033 per 1K tok (cloud) &
Oct-2025 & Microsoft AI Foundry pricing [21] & Estimated \\
\end{longtable}
}

Notes: ``Estimated'' entries derive from provider sheets plus token
overhead assumptions; Humains-Junior uses measured throughput and rental
rates converted to per-1K tokens, then adjusted for utilization/batch
(see E.3).

\subsection{Appendix F: Exploratory Optimized-Prompt Frontier Results (Non-Comparable)}\label{appendix-f-exploratory-optimizedprompt-frontier-results-noncomparable}

\subsection{Context and Rationale}\label{f.1-context-and-rationale}

To explore upper bounds with customized prompting, we evaluated
model-specific optimized prompts (3--5 few-shot examples) under an
earlier judge configuration (Gemini 1.5 Pro, GPT-4o, Claude 3.5). These
settings differ from the unified prompt and updated judges used for the
main study, and are therefore not directly comparable. We report them
for completeness.

\subsection{Results Summary (Exploratory)}\label{f.2-results-summary-exploratory}

{\renewcommand{\LTcaptype}{} 
\begin{longtable}[]{@{}
  >{\raggedright\arraybackslash}p{(\linewidth - 8\tabcolsep) * \real{0.2125}}
  >{\raggedright\arraybackslash}p{(\linewidth - 8\tabcolsep) * \real{0.1000}}
  >{\raggedright\arraybackslash}p{(\linewidth - 8\tabcolsep) * \real{0.2250}}
  >{\raggedright\arraybackslash}p{(\linewidth - 8\tabcolsep) * \real{0.1875}}
  >{\raggedright\arraybackslash}p{(\linewidth - 8\tabcolsep) * \real{0.2750}}@{}}
\toprule\noalign{}
\begin{minipage}[b]{\linewidth}\raggedright
Model
\end{minipage} & \begin{minipage}[b]{\linewidth}\raggedright
Baseline
\end{minipage} & \begin{minipage}[b]{\linewidth}\raggedright
Optimized Scaffold
\end{minipage} & \begin{minipage}[b]{\linewidth}\raggedright
Few-Shot (opt.)
\end{minipage} & \begin{minipage}[b]{\linewidth}\raggedright
Judge Config (earlier)
\end{minipage} \\
\midrule\noalign{}
\endhead
\bottomrule\noalign{}
\endlastfoot
GPT-4o & 78\% & \textbf{91\%} & 3--5 & Gemini 1.5, GPT-4o, Claude 3.5 \\
Claude 3.5 Sonnet & 83.3\% & \textbf{90\%} & 3--5 & Gemini 1.5, GPT-4o,
Claude 3.5 \\
\end{longtable}
}

Notes: Different judge panel (earlier) and few-shot counts for the
entries above; use only as illustrative upper bounds. Main results use a
unified prompt and updated judges (Gemini 2.5 Pro, GPT-4o, Claude 3.7).

\end{document}